%% file: main.tex
\documentclass[10pt,twocolumn,letterpaper]{article}

\usepackage[pagenumbers]{cvpr} 
\usepackage{soul}
\usepackage[dvipsnames]{xcolor}

\input{preamble}

\definecolor{cvprblue}{rgb}{0.21,0.49,0.74}
\usepackage[pagebackref,breaklinks,colorlinks,citecolor=cvprblue]{hyperref}

\def\paperID{3787} 
\def\confName{CVPR}
\def\confYear{2024}

\title{\textit{SHARE}: {\it S}ingle-view {\it H}uman {\it A}dversarial {\it RE}construction}

\author{Shreelekha Revankar\\
{\tt\small revankar@umd.edu}
\and
Shijia Liao\\
{\tt\small lengyue@terpmail.umd.edu}
\and
Yu Shen\\
{\tt\small yushen@umd.edu}
\and
Junbang Liang\\
{\tt\small junbangl@amazon.com}
\and 
Huaishu Peng\\
{\tt\small huaishu@umd.edu}
\and
Ming Lin\\
{\tt\small lin@umd.edu}
}
\begin{document}
\maketitle
\input{sec/0_abstract}    
\input{sec/1_intro}

\input{sec/2_relatedwks}
\input{sec/3_method}
\input{sec/4_exp}
\input{sec/5_conclusion}
{
    \small
    \bibliographystyle{ieeenat_fullname}
    \bibliography{main}
}

\input{sec/X_suppl}

\end{document}

%% file: preamble.tex
\usepackage[dvipsnames]{xcolor}


%% file: sec/0_abstract.tex
\begin{abstract}
The accuracy of 3D Human Pose and Shape reconstruction (HPS) from an image is progressively improving. Yet, no known method is robust across all image distortion. To address issues due to variations of camera poses, we introduce \textit{SHARE}, a novel fine-tuning method that utilizes adversarial data augmentation to enhance the robustness of existing HPS techniques. 
We perform a comprehensive analysis on the impact of camera poses on HPS reconstruction outcomes. We first generated large-scale image datasets captured systematically from diverse camera perspectives. We then established a mapping between camera poses and reconstruction errors as a continuous function that characterizes the relationship between camera poses and HPS quality. Leveraging this representation, we introduce \textbf{RoME (Regions of Maximal Error)}, a novel sampling technique for our adversarial fine-tuning method. 

The SHARE framework is generalizable across various single-view HPS methods and we demonstrate its performance on HMR, SPIN, PARE, CLIFF and ExPose. Our results illustrate a reduction in mean joint errors across single-view HPS techniques, for images captured from multiple camera positions without compromising their baseline performance. In many challenging cases, our method surpasses the performance of existing models, highlighting its practical significance for diverse real-world applications.
\end{abstract}

%% file: sec/1_intro.tex
\section{Introduction}

The reconstruction of human body pose and shape (HPS) has gained attention from industries like fashion, healthcare, special effects, surveillance, computer animation, and virtual and augmented reality~\cite{Liang01,hu2018personalized,ghadi2021syntactic}. Single-view 3D human pose and shape recovery is of particular interest due to its simplicity and practicality, sparking renewed research efforts to enhance its accuracy and robustness.

\input{figures/pipeline_intro}
Common issues affecting HPS reconstruction results, such as self-occlusion, low-contrast lighting or poor depth perception are often due to suboptimal camera poses~\cite{Puscas01,kocabas2021spec,pang2022benchmarking}.
Therefore, it is vital to understand how camera poses impact reconstruction quality. Furthermore, in large-scale consumer applications of HPS, such as virtual try-on or healthcare, these effects can significantly influence the user experience.

In this paper, we study the influence of various camera poses on HPS using images. We propose novel methods to minimize disparities due to such camera pose variations. 
Our adversarial fine-tuning method complements numerous pre-trained HPS models and improves their robustness against diverse camera poses not commonly found in their training data.

The key contributions of this work include:
\begin{enumerate}
\item A framework for the automatic creation of large-scale image datasets for given bodies, camera poses, and scene settings (Sec.~\ref{subsection:GenSyn}) to be publicly released;
\item A systematic study and analysis on the {\em impact of camera poses} on the quality of human pose and shape reconstruction (Sec.~\ref{subsection:AnalysisOfPose});
\item An {\em adversarial data augmentation} technique for fine-tuning pre-trained HPS models against image variation due to camera poses using {\em differentiable sampling techniques} (Sec.~\ref{subsection:AdvDataAug});
\end{enumerate}

%% file: figures/pipeline_intro.tex
\begin{figure}[htpb]
    \centering
    \includegraphics[width=\columnwidth]{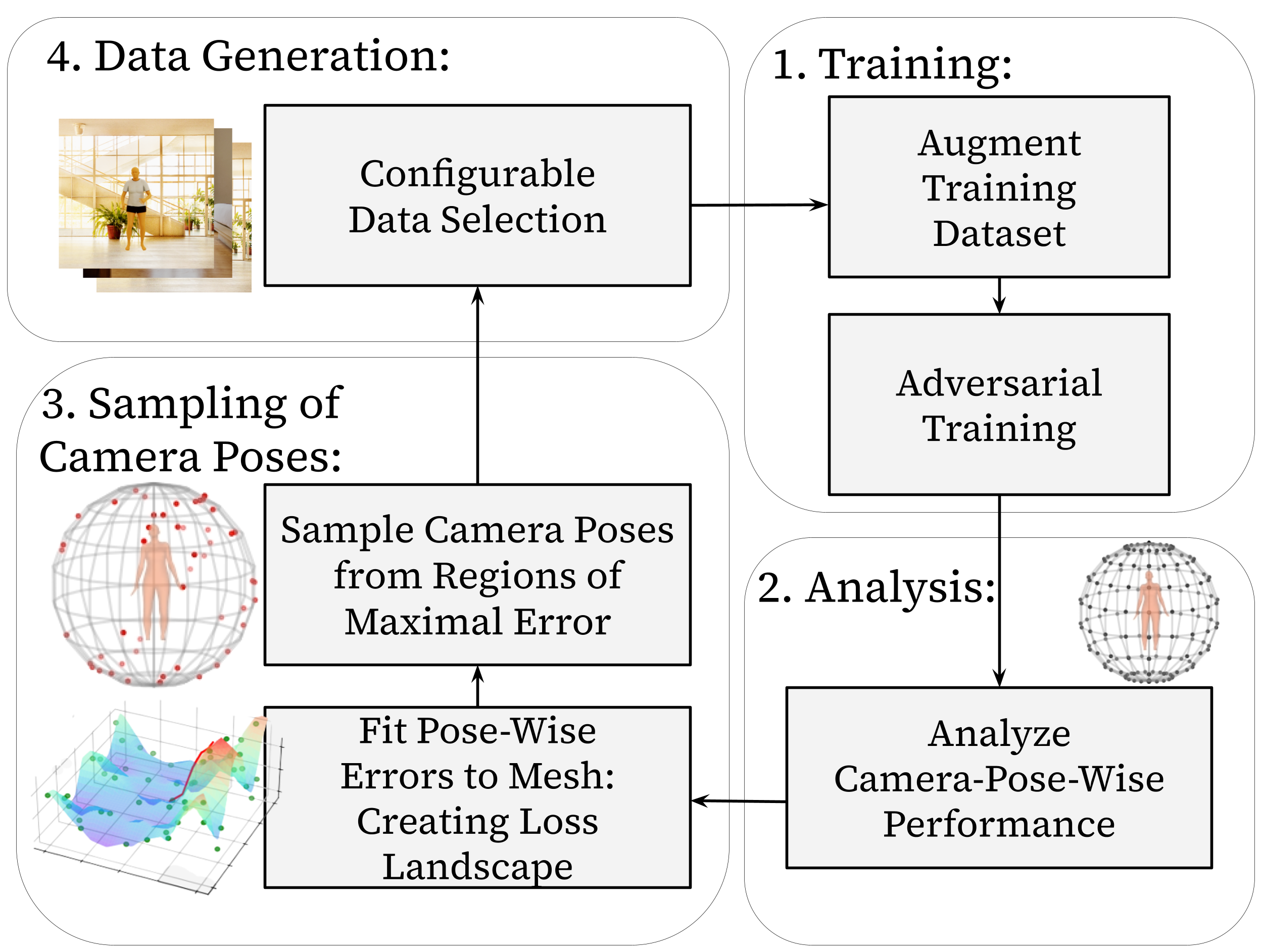}
    \vspace*{-1.5em}
    \caption{{\bf The SHARE Framework} adversarially augments and modifies synthetic training data for a single-view HPS model. It is initialized by generating training data from all camera poses. Each iteration of SHARE operates in four phases: (1) augmenting the model's training data to train the model, (2) assessing camera-pose-wise performance, (3) sampling the most adversarial camera poses and (4) down-selecting a new training dataset for augmentation using RoME sampled poses.}
    \vspace{-1.0em}
    \label{fig:pipeline}
\end{figure}

%% file: sec/2_relatedwks.tex
\section{Related and Concurrent Works}

We provide a review of recent advancements in human pose and shape estimation (HPS), along with the availability of datasets specifically curated for this purpose. We also explore the challenges posed by image distortions, specifically focusing on those caused by camera pose variation, and discuss the application of different techniques to improve the robustness of HPS.

\subsection{Data-driven methods for HPS}
The process of reconstructing human pose and shape through data-driven methods typically involves the application of machine learning techniques to establish a correlation between 2D images and 3D body models.

While techniques utilizing multiple viewpoints, video footage, or a combination of visual and other sensory inputs have shown enhanced reconstruction capabilities over single-view or monocular methods~\cite{Liang01,Sengupta02,Zins01,hu20213dbodynet,lu2018accurate,ChengKinect}, the demand for robust single-view reconstruction remains paramount in many large-scale use cases of HPS~\cite{hmrKanazawa17,Federica01,Omran01,zheng2019deephuman,pavlakos2019expressive,zanfir2021neural,choutas2020monocular,xiang2019monocular}, especially in real-world applications like virtual try-on.

At the heart of these data-driven approaches lies the existence of extensive datasets, comprising of human images sourced from various outlets, including online images, motion capture sessions, and artificially generated images based on 3D body models~\cite{kocabas2021pare,hmrKanazawa17,kolotouros2019spin,li2022cliff}. Despite the remarkable progress achieved by these methods, it is crucial to acknowledge a significant challenge: the limitations of training data.

\subsection{Datasets for HPS}
The data behind data-driven HPS models has evolved over time.
Early datasets, laid the foundation for such an approach~\cite{elhayek2015efficient,elhayek2016marconi,liu2011markerless}. More recent datasets have emerged with distinct objectives; however, they often lack 3D ground truth annotations~\cite{lin2014microsoft,Johnson10leeds,andriluka14cvprmpii,zhang2019pose2seg}.

In response to this, recent efforts have involved fitting body models to these datasets to derive new 3D "ground truth" information for 2D human images~\cite{joo2020exemplar,sengupta2020synthetic,leroy2020smply,zhang2020object}. 


Obtaining accurate 3D ground truth data for evaluating reconstruction methods is challenging and time-consuming in the real world, leading real-world datasets to have varying limitations. Some require full-body motion capture~\cite{loper2014mosh,sigal2010humaneva}, which restricts clothing variety or necessitate a controlled lab environment~\cite{joo2015panoptic,yu2020humbi}. Others employ marker-less motion capture with Inertial Measurement Unit (IMU) sensors. 
Thus, inherently, this ground truth information is susceptible to measurement uncertainties and sensor errors, irrespective of the collection method or source. 
Despite these challenges, such datasets provide the most realistic inputs for our models~\cite{sigal2010humaneva,Ionescu01,trumble2017total,mehta2018single}. Datasets like 3DPW~\cite{vonMarcard20183dpw} and MPI-INF-3DHP~\cite{mono-3dhp2017} are favored for HPS evaluation due to their mobile nature and multi-view capabilities. 

Nevertheless, for our goal of analyzing the effect of camera poses variations on HPS, real-world datasets typically lack the necessary comprehensive information on camera poses and camera details during image capture.




Advancements in computer graphics have facilitated the creation of synthetic or simulated datasets that provide known ground truth details, including body sizes, shapes, and poses, which are often absent in real-world datasets~\cite{Patel:CVPR:2021,Black_CVPR_2023,mehta2018single,pumarola20193dpeople,ranjan2020learning,varol2017learning,mono-3dhp2017}


While some may consider these generated datasets as not realistic enough for human eyes, for image processing (where edges, features, and patterns are critical for machine perception) and neural network training (when there is insufficient data to represent corner cases), the use of simulated data is perhaps one of the best practical alternatives, as proven in many recent works on HPS and otherwise~\cite{varol2017learning,dosovitskiy2015flownet}. 

To generate such data, 3D human body representations have been widely employed~\cite{SMPL:2015,chen2019parametric,reed2014developing,briceno2018makehuman,adobe2020mixamo,Renderpeople01,humano3d}. Simulated data has demonstrated its potential to enhance the accuracy of HPS methods, improving reconstruction results, as indicated in various studies~\cite{chen2016synthesizing, Liang01}. However, datasets which offer diverse bodies and rendering settings, often have limited camera pose ranges and may lack publicly available human models, making it difficult to capture additional data from new camera perspectives~\cite{Patel:CVPR:2021,pumarola20193dpeople,Black_CVPR_2023,varol2017learning}. This highlights \textbf{the need for a dataset that comprehensively covers a wide range of camera perspectives.} 

\subsection{Robustness \& Adversarial Data Augmentation}

Several studies have highlighted the presence of failures in HPS caused by challenging depth ambiguities~\cite{hmrKanazawa17,Federica01,Omran01}. It is recognized that factors like camera pose and self-occlusion significantly impact depth perception~\cite{pang2022benchmarking,todd2007effects}. However, the specific influence of a camera pose on reconstruction results has not been extensively investigated in the existing literature.

Researchers have introduced adversarial techniques~\cite{Wandt2019RepNetWS,chou2018self,ke2011view,hmrKanazawa17} and regression networks~\cite{jackson20183d,chen20223d,guler2019holopose,kocabas2021pare, zhang2021pymaf, zhang2023pymaf} to tackle (self-)occlusions and improve the overall quality of HPS.

Liu et. al, Sun et. al., and Sardari et. al among others have also worked on creating camera pose invariant methods for HPS, but these works present entirely new reconstruction paradigms.~\cite{Sun01,liu2021view,sardari2021unsupervised}.
Other techniques, such as CanonPose~\cite{wandt2021canonpose}, AdaptPose~\cite{gholami2022adaptpose}, and SPEC~\cite{kocabas2021spec}, focus on accurately predicting camera poses from images. However, these inverse techniques often require additional components, such as training separate models specifically for inferring camera parameters. Moreover, in the case of SPEC, the training details are not yet available. In contrast, \textbf{our objective is to develop a generalizable approach that directly improves the robustness of existing and future HPS models themselves.}

Outside of HPS, in the field of machine learning, efforts have been made to address biases and improve the robustness of datasets and models. Techniques include introducing corruptions or biases to existing datasets to evaluate neural network robustness~\cite{hendrycks2018benchmarking}. These datasets can also be utilized for adversarial machine learning, a method that enhances model robustness. Adversarial data augmentation is a notable technique in this regard.

Ghosh et al. analyze the impact of quality degradations on convolutional neural networks, leading to improved learning outcomes~\cite{ghosh-robustness}. Cubuk et al. propose a method for searching enhanced data augmentation policies~\cite{cubuk2018autoaugment}. Various frameworks have demonstrated that adversarial data aumentation can enhance model robustness~\cite{hendrycks2019augmix,devries2017cutout,zhang2018mixup,yun2019cutmix}. Shen et al. use adversarial training to improve robustness in autonomous driving by addressing different perturbations~\cite{shen2021gradient}. These works provide valuable insights that can inspire advancements in HPS.

\textbf{In this work, we present a new adversarial, fine-tuning framework for human pose and shape regression models, to improve their robustness against camera pose variation.} In contrast to most recent related works~\cite{hendrycks2019augmix,shen2021improving}, SHARE analyzes specific camera pose variations with respect to the human figure and focuses on the impact of camera perspectives that lead to poor performance on 3D HPS, thereby \textbf{hardening existing HPS methods without retraining an entirely new model}.

%% file: sec/3_method.tex
\section{Methodology}

Our aim is to improve robustness in human pose and shape recovery against ubiquitous perturbations caused by camera pose variation. We begin with the large-scale generation of data (Sec.~\ref{subsection:GenSyn}) for sensitivity analysis of pre-existing HPS methods (Sec.~\ref{subsection:AnalysisOfPose}). Through this systematic sensitivity analysis, we can approximate to what degree a camera pose may affect reconstruction (Sec.~\ref{generation_of_LL}). With these results we implement an adversarial framework \textit{SHARE} to rectify the disparities created by camera pose variation (Sec.~\ref{subsection:AdvDataAug}). We illustrate this method in Fig.~\ref{fig:pipeline}.

\subsection{Data Generation}
\label{subsection:GenSyn}

\input{figures/synthetic_examples}
We introduce an automated human image dataset generator that relies on a rendering engine, a human body model, and efficient configuration.

To create and configure human bodies in our generator, we incorporated body models from RenderPeople~\cite{Renderpeople01} and the Skinned Multi-Person Linear Model (SMPL). The SMPL model offers extensive control over body shapes, sizes, and poses, with 82 SMPL parameters~\cite{SMPL:2015}. We obtained a wide range of realistic human poses from large-scale datasets~\cite{Ionescu01, sigal2010humaneva} to complement the body model. This combination enabled the generation of diverse bodies for use in our rendering environment.

To enable our automatic generator's functionality, we developed a script capable of receiving user-specified parameters and settings for rendering. These parameters encompass a wide array of features, such as human body proportions, poses, clothing, and skin tone, as well as rendering factors like lighting, background, and camera positions.

In response to these requirements, the script dynamically generates a configuration file that is compatible with multiple rendering environments. Our image generator has been tested on both Unity~\cite{juliani2018unity} and Blender~\cite{Blender}, with scripts designed to accept various configuration files and produce rendered images, complete with relevant details.

The fusion of rendered images, corresponding camera perspectives, and ground truth body parameters derived from the generated human body models forms a comprehensive human image dataset suitable for tasks like human pose and shape reconstruction. \textbf{This generator will be publicly released.}

Employing our image generator, we produced multiple images from 2500 viewpoints encircling the body models within a polar coordinate system. Fig.~\ref{fig:synthetic_images} exhibits select images from our dataset, highlighting the distinctive camera perspectives. This dataset serves as the basis for a sensitivity analysis to examine the influence of camera poses.

\input{figures/solo_cyclical}

\subsection{Analysis of Camera Poses}
\label{subsection:AnalysisOfPose}

Common distortions such as self-occlusion or low-contrast can often be the result of poor camera positioning~\cite{Puscas01,kocabas2021spec,pang2022benchmarking}. Therefore, our objective, after developing a dataset generator, was to evaluate the impact of different camera positions. We aimed to determine whether specific camera poses resulted in improved body reconstruction outcomes and to identify those that led to suboptimal results across a wide range of human bodies and environmental settings

To evaluate the influence of different camera poses, we created a dataset consisting of 10,000 images, uniformly sampled from 2,500 distinct camera perspectives. Each image portrays a unique human body with randomly chosen body poses, clothing, skin tones, lighting conditions, and environmental settings. This dataset is referred to as our evaluation dataset (All Camera Poses) {\em to be publicly released.}

We selected four pre-trained HPS models as our baselines from the OpenMMLab 3D Human Parametric Model Toolbox and Benchmarks, which are available under the Apache License 2.0~\cite{mmhuman3d}to maintain consistency and replicatability. We chose Human Mesh Recovery (HMR)~\cite{hmrKanazawa17}, SMPL with optimization IN the loop (SPIN)~\cite{kolotouros2019spin}, Part Attention REgressor (PARE)~\cite{kocabas2021pare} and Carrying Location Information in Full Frames (CLIFF)~\cite{li2022cliff}. 

We employed our evaluation dataset and our baseline models to reconstruct the simulated bodies and computed the average errors for every camera pose. Our metrics were the standard mean per joint position error (MPJPE) metric to evaluate reconstructed human bodies, as well as a variation that includes Procrustes alignment (PA-MPJPE)~\cite{Ionescu01}. 

Looking specifically at a current state-of-art model, PARE, the average PA-MPJPE from all camera poses 124.28 mm, which is significantly greater than the reported average PA-MPJPE of 50.78mm on another popular evaluation dataset 3DPW~\cite{vonMarcard20183dpw}. This difference indicates that the larger variety of camera poses creates a great impact on reconstruction results. Thus, rectifying any losses due to camera pose variations can improve reconstruction accuracy.

Upon plotting the camera poses against their associated errors (Fig.~\ref{subsection:AnalysisOfPose}), \textbf{we observed that specific regions consistently exhibited better or worse performance regardless of body poses}. To validate this observation, we generated two additional datasets, each featuring a singular distinct body in a distinct pose. We then calculated the average error across each of the 2500 camera positions. We found that while the error values and variances differed, \textbf{the overall error patterns remained consistent}. These results can be found in the appendix.

We could now readily distinguish the camera pose regions that excelled and those that underperformed. Notably, \textbf{as the camera perspectives shifted towards the front side of the human body, we noticed substantial reductions in errors, indicating an improved quality of reconstruction. Additionally, when transitioning from a higher to a lower camera perspective, a distinct and recurring error pattern emerged.}

The cyclical nature of the performance across camera poses demonstrated using PARE can be noted in Fig.~\ref{fig:cyclical_errors}. While Fig.~\ref{fig:cyclical_errors} plots the average errors across images from camera poses in the evaluation dataset, which contains a diverse set of bodies and poses. We include similar plots with the error curves for a singular body/pose as well as the other baselines in the appendix.

Upon inspecting the images from different regions, \textbf{we found that the camera perspectives with the lowest errors were those captured near or slightly below the waistlines in the front view.} These angles displayed enhanced depth perception and reduced self-occlusion. In contrast, images taken from a top-down, peering angle yielded the poorest reconstruction quality. This can be attributed to limited depth perception and occlusion due to the chest or head obstructing other body parts.

This discovery holds significance in the context of deploying image-based HPS models since the camera poses with suboptimal performance align with what are commonly referred to as "selfie" angles~\cite{eckel2018selfie}.

As real-world images are not curated by experts specifically for training HPS methods, they may not adhere to the same criteria as the images used in training and testing. To ensure the reliability of user input in HPS, the reconstruction quality must be resilient to limited depth perception resulting from variations in camera perspective.

\subsection{Camera Poses and Reconstruction Errors}
\label{generation_of_LL}

Leveraging the consistent behavior of camera poses on reconstruction results across a wide variety of bodies and poses, we can create a mapping between relative camera perspective and reconstruction error. Providing us with a continous representation of the relationship between relative camera poses and their predicted reconstruction results.

Such a continous representation can be very useful in real-world scenarios where ground truth information is not easy to obtain.

We model human pose and shape estimation as \(G\), which accepts an image and approximates human mesh parameters,
consisting of \textit{shape}, \(\beta\), and \textit{pose}, \(\alpha\).

Let \(p\) be an image captured of a human. \(p\) can be characterized by the camera perspective in polar coordinates (\(\theta,\phi\)) relative to the human \(h\). 
We define the ground truth characteristics of \(h\) as \(h(\beta_{gt},\alpha_{gt} )\) and
\(p\) can be expressed as \(p((\theta,\phi),h)\). Giving us the following:
\begin{equation}\label{eq:jointreconstruction}
    G(p((\theta,\phi),h)) = \{\beta,\alpha\}
\end{equation}
\noindent
These parameters can be utilized by a parametric model~\cite{SMPL:2015, chen2019parametric, reed2014developing} to generate human meshes.

The general goal of most HPS regression models is to accurately reduce the difference between the reconstructed body joints and the joints of the human. In other words, their goal is to minimize the loss \(L_{3d}\). 
Commonly this loss is calculated as such:
\begin{equation}\label{eq:loss}
    L_{3d} = ||\alpha_{gt}-T(\alpha)||^{2}_{2}
\end{equation}
where \(\alpha_{gt}\) represents true joints of the human pose to be reconstructed and $T$ is rigid transformation $T(x) = s$ x $R + t$ for Procrustes Alignment. With $t,s,R$ as the translation, rotation matrix and scaling factor for Procrustes Alignment respectively. The full details of this transformation is included in the appendices.

Using Eqn.~\ref{eq:jointreconstruction} and ~\ref{eq:loss}, we obtain:
\begin{equation}\label{eq:ourtechnique}
    L_{3d} = ||\alpha_{gt}-G(p((\theta,\phi),h))||^{2}_{2} = f(\theta,\phi)
\end{equation}

By sampling over $\theta$ and $\phi$, we are able to recover the continuous representation of $L_{3d} = f(\theta,\phi)$ in a numerical way, then use an neural network to approximate $f$. After that, we can use partial derivatives \(\frac{df}{d\theta}\) and \(\frac{df}{d\phi}\) to describe the sensitivity of \(L_{3d}\) w.r.t. \(\theta\) and \(\phi\), thus modelling the relationship between relative camera poses and reconstruction errors.

\subsubsection{Generation of Loss Landscape}
\input{figures/loss_landscape}
With the ability to generate a large dataset of images from a vast number of camera poses, we can fit camera-pose-wise errors to a mesh to create a loss landscape that can be used to understand the impact of a change in camera poses for a specific model.

Since the camera positions are in spherical coordinates, and our radius is fixed for the evaluation dataset, we can use the $\theta$ and $\phi$ of the camera position as the $x$ and $y$ coordinates of each camera pose in our mesh.

We then train a multi-layer perceptron to predict the error given a $\theta$ and $\phi$ location, following the principle described in eq.~\ref{eq:ourtechnique}. 
This approximated error $E$ is used along the $z$ axis for each camera pose. We then scale the $E$, $\theta$ and $\phi$ into the range [-1, 1] using min-max feature scaling, as seen for PARE in Fig.~\ref{fig:loss_landscape}.

The large variance in results from the different camera poses proves that there is a need to ameliorate HPS methods against camera pose variation. We provide a technique to do so, \textit{SHARE}, a major contribution of our paper. SHARE is an adversarial data augmentation fine-tuning technique to make existing HPS models more robust to camera pose variation (Sec.~\ref{subsection:AdvDataAug}). A camera pose-wise situated loss landscape is an important backbone to sampling methods employed in \textbf{SHARE} and can allow us to visualize the predicted performance of camera poses in relation to one another. 


\subsection{SHARE Adversarial Data Augmentation}
\label{subsection:AdvDataAug}
Our method, Single-view Human Adversarial Reconstruction (SHARE), serves as an adversarial fine-tuning technique applicable to a wide range of pre-trained HPS models.

The SHARE framework operates by adversarially augmenting and modifying training data for a single-view HPS model at specific intervals.

The process commences with the generation of training and validation data using our data generator {\em from all camera poses}, described in Section~\ref{subsection:GenSyn}. The validation dataset comprises of images from various camera poses, each paired with its expected output.

SHARE operates in four phases within each interval. (See Fig.~\ref{fig:pipeline}):

\textbf{(1)} Firstly, we augment a small percentage of the model's original training data with our training data, then train the model using its native training paradigm for a predefined number of epochs.

\textbf{(2)} Next, we assess the model's performance using our validation dataset to compute example-wise, i.e., camera-pose-wise errors. These results enable us to construct a loss landscape, as described in Sec.~\ref{generation_of_LL}. 

\textbf{(3)} With this continuous representation of camera-pose-wise performance, we employ a sampling technique to select camera poses associated with higher errors.

\textbf{(4)} Subsequently, we generate a new training dataset for adversarial data augmentation using the sampled camera poses, and the next interval commences.

\textbf{Data Diversity} During dataset generation, we enhance human body and pose diversity by randomly selecting shape/pose parameters~\cite{Federica01} from the MPII dataset~\cite{andriluka14cvprmpii}, which is based on real-world bodies and poses. We also randomly sample background environments, lighting conditions, skin tones, body sizing, and clothing.

\textbf{Compatibility with Existing Models}
SHARE seamlessly integrates with various existing single-view HPS regressors, provided they adhere to similar input-output formats—commonly employed by state-of-the-art methods. This inherent compatibility underscores SHARE's \textit{generalizability} and its applicability to enhance a wide range of approaches. 
\subsubsection{Sampling Techniques}
Within the context of the SHARE framework, the 3rd phase includes sampling camera poses from our loss landscape. 
The most direct approach is the \textbf{Greedy} method, which entails selecting the worst performing camera positions. One way to implement this is by choosing all camera positions that produce errors higher than the mean error and then sampling new images from these poses. However, this approach treats all camera poses performing worse than average, whether slightly or significantly, in the same manner.

For better performance, we propose a \textbf{Regions of Maximal Error} (RoME) sampling method. This sampling technique ensures that the regions with the highest error are sampled more densely than others. {\bf RoME Sampling} works on the idea that we can assign a local average to a {\em region} on our loss landscape.

Suppose that we have $N$ samples in the form of $(\theta,\phi, E)$ obtained through evaluating on our validation dataset. We generate a loss landscape $f(N)$ to create a continuous representation of the relationship between camera poses and their reconstruction errors~(\ref{generation_of_LL}). 

Our samples are now defined as $\{(X_n, Y_n, E_n)\}_{n=1}^{N}$, 
where $X_n, Y_n$ are coordinates of the sample on our landscape, and $E$ represents the predicted error along the $z$ axis.

At each sample point, we calculate the first $f'_{E,n}$ and second $f''_{E,n}$ derivatives. By combining the error with its first and second derivatives, we create a composite metric $W_n$ that accounts for both the magnitude of the error and the slope and curvature of the loss landscape at that point: $W_n = |E_n| + f'_{E,n} + f''_{E,n}$.
As a result, our samples are now defined as $\{(X_n, Y_n, W_n)\}_{n=1}^{N}$.

We define a variable $P$ to denote the number of partitions we wish to create within our mesh. For instance, if $P=8$ we create $8^3 = 512$ regions within our mesh. 

We calculate a threshold $\tau$ which is computed as the mean of all $W_n$ in our loss landscape and remove regions that do not contain any samples surpassing $\tau$ from our sampling pool.

With the remaining regions, we calculate the mean $W_n$ for all samples within a region. These average regional scores are denoted by the set $\{AW_r\}^R_{r=1}$.

For regions where the average regional score is greater than the threshold, i.e. $AW_r > \tau$, we employ a random selection process. $M$ samples are randomly selected from such regions, ensuring a degree of diversity in the chosen samples, where $M = \alpha N/P^3$, where $\alpha$ is the scaling factor determining the proportion of the total available samples in $N$ to be selected based on the parameter $P$.

Conversely, in regions where the average region score is below or equal to the threshold, i.e. $AW_r \leq \tau$, we opt to select only $M/\alpha$ sample from these regions.

\textbf{The selected samples from RoME sampling excel in denser sampling from regions with poorer performance, optimizing both time and space efficiency compared to indiscriminate methods (e.g. random sampling).} These samples are then used to generate new adversarial examples for the next iteration of SHARE.

%% file: figures/synthetic_examples.tex
\begin{figure}[t]
    \centering
    \includegraphics[width=0.24\columnwidth]{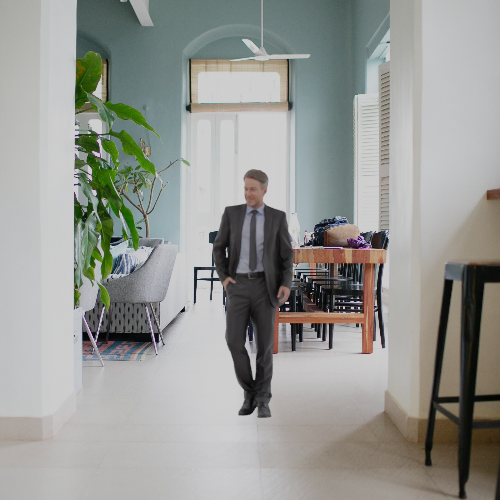}
    \includegraphics[width=0.24\columnwidth]{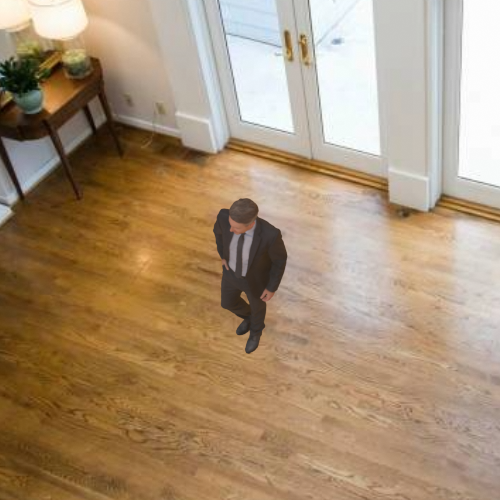}
    \includegraphics[width=0.24\columnwidth]{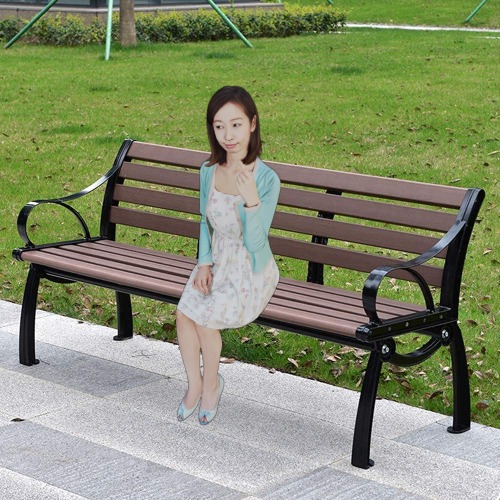}
    \includegraphics[width=0.24\columnwidth]{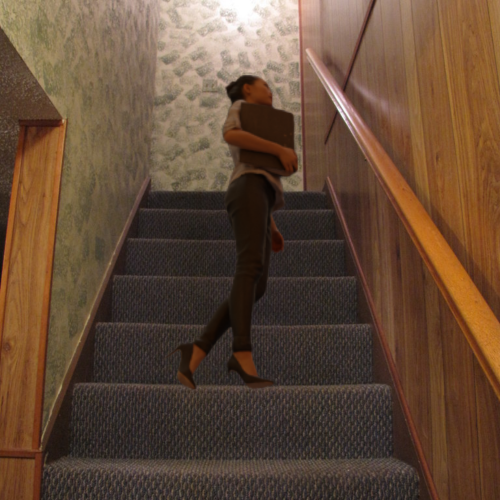}
    \includegraphics[width=0.24\columnwidth]{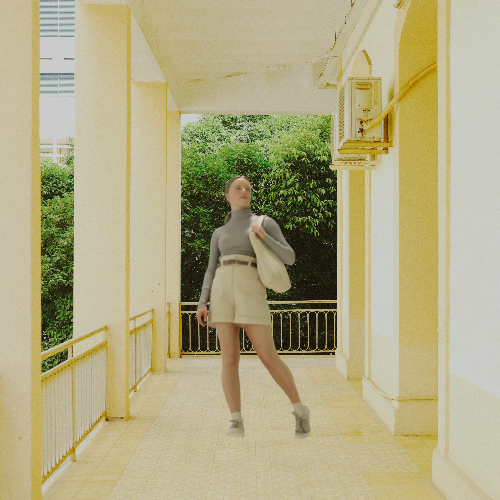}
    \includegraphics[width=0.24\columnwidth]{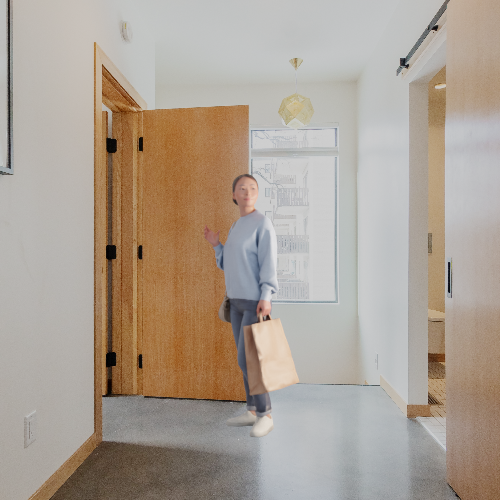}
    \includegraphics[width=0.24\columnwidth]{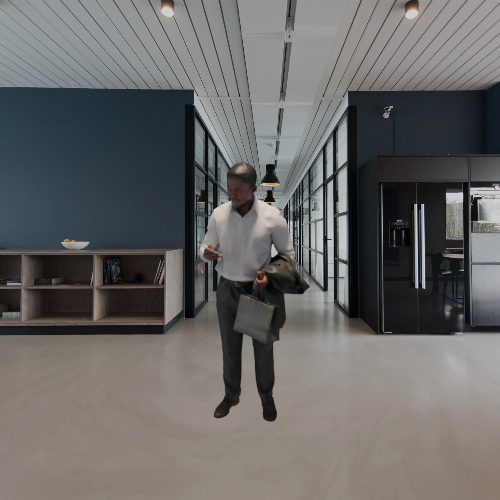}
    \includegraphics[width=0.24\columnwidth]{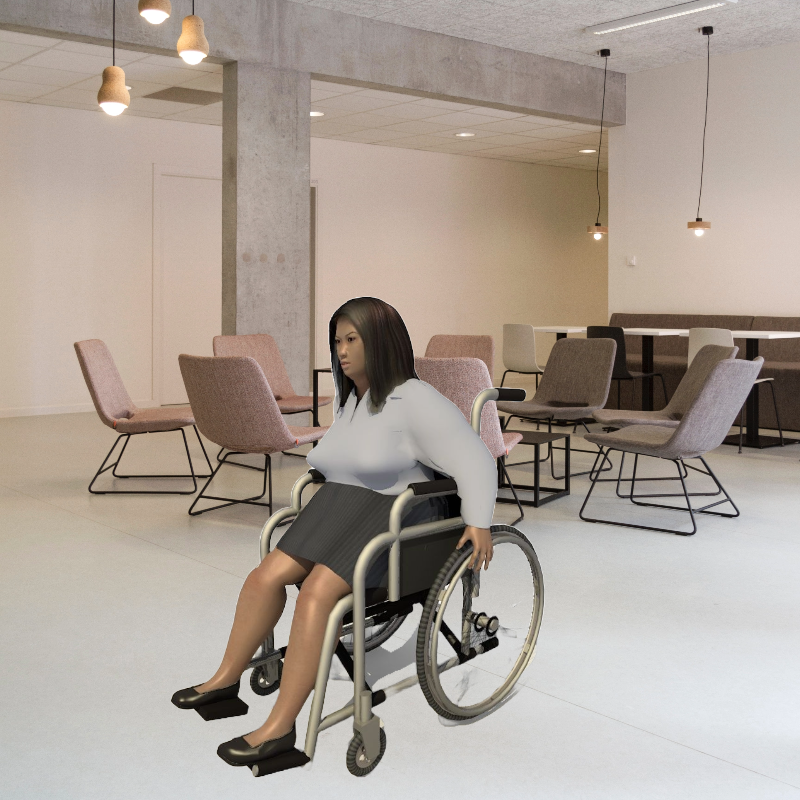}
    
    \caption{Examples of images generated using our data generator from various camera poses.}
    \label{fig:synthetic_images}
\end{figure}

%% file: figures/solo_cyclical.tex
\begin{figure*}[t]
    \centering
    \vspace*{-0.5em}
    \includegraphics[width=0.7\textwidth]{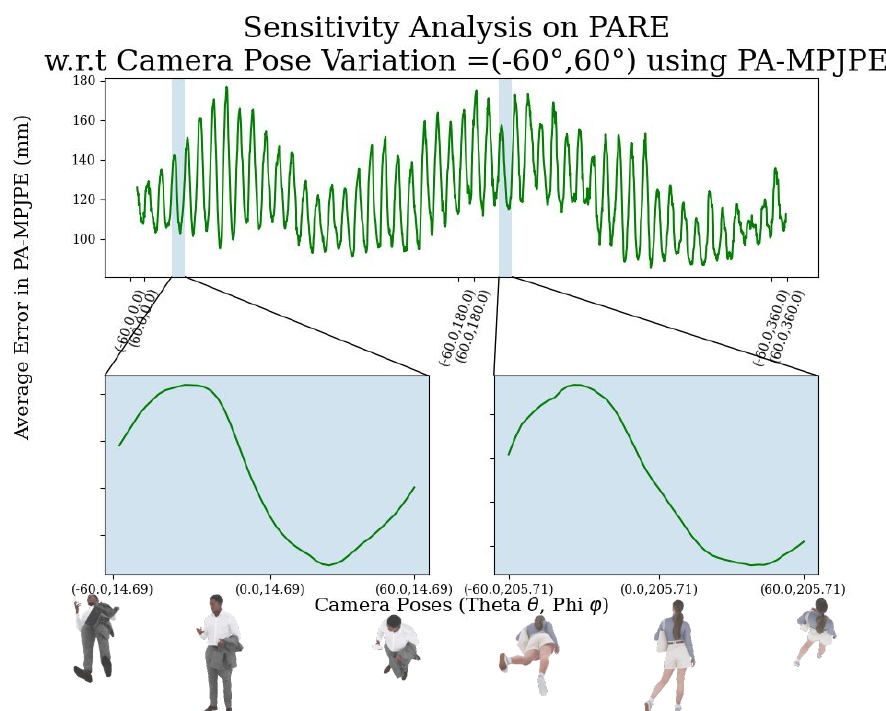}
    \vspace*{-1em}
    \caption{\textbf{Sensitivity analysis on PARE~\cite{kocabas2021pare} with respect to camera pose using PA-MPJPE.} The x-axis iterates through all camera poses $(\theta,\phi)$, where $\phi$ represents the azimuthal angle around the body (0, 360), and $\theta$ represents the vertical viewing angle (-60, 60) for each $\phi$. The y-axis represents the average error in PA-MPJPE over a diverse dataset encompassing a wide range of bodies, body poses, and environments. This plot explicitly depicts the average error associated with each camera pose, revealing \textbf{a discernible oscillatory bias with varying performance in different regions around the human body.} Additional plots for single-person datasets and comparisons with other HPS techniques are available in the appendix.}
    \vspace{-1.em}
\label{fig:cyclical_errors}
\end{figure*}

%% file: figures/loss_landscape.tex
\begin{figure}[t]
    \centering
    \vspace{-1.5em}
\includegraphics{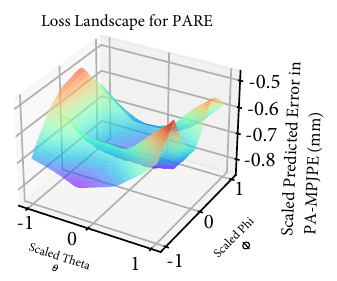}
    \caption{\textbf{Loss Landscape for PARE~\cite{kocabas2021pare}}. The x-axis represents the scaled $\theta$ values, while the y-axis represents the scaled $\phi$ values, the z-axis depicts the predicted PA-MPJPE associated with a given camera pose. We include the loss landscapes for all baselines in the appendix.}
    \label{fig:loss_landscape}
    \vspace{-1.em}
\end{figure}

%% file: sec/4_exp.tex
\section{Experimental Setup}
SHARE is applicable to common HPS regressors that infer parametric body models ~\cite{SMPL:2015,chen2019parametric,reed2014developing}. We illustrate its benefit using the following methods: HMR~\cite{hmrKanazawa17}, SPIN~\cite{kolotouros2019spin}, PARE~\cite{kocabas2021pare} and CLIFF~\cite{li2022cliff}. We further demonstrate the extension of SHARE into body parts using ExPose-hand~\cite{choutas2020monocular}.

\textbf{Training:}
For HMR and SPIN, the training data was composed of a mixture of the Moshed Human3.6M dataset~\cite{Ionescu01}, COCO~\cite{lin2014microsoft}, MPI-INF-3DHP~\cite{mono-3dhp2017}, LSP~\cite{Johnson10leeds}, LSPET~\cite{Johnson10leeds}, and MPII~\cite{andriluka14cvprmpii}. For PARE the training data was comprised of the Moshed Human3.6M dataset, MPI-INF-3DHP, EFT-COCO, EFT-LSPET, and EFT-MPII~\cite{joo2020exemplar}. For CLIFF the training data was composed of the Moshed Human3.6M, MPI-INF-3DHP, COCO and MPII with pseudo-GT provided by the CLIFF annotator for the latter two. ExPose-hand was trained using the FreiHand~\cite{zimmermann2019freihand} dataset.

To ensure consistency and reproducibility, we trained baseline models using MMHuman3D~\cite{mmhuman3d}, an open-source computer vision platform developed by OpenMMlab. We maintain identical training data and schemes as the original HPS models for fair comparisons across benchmark settings.

\textbf{Evaluation \& Metrics:}
We evaluate quantitatively on the test sets of 3DPW~\cite{vonMarcard20183dpw}, MPI-INF-3DHP~\cite{mono-3dhp2017}, FreiHand~\cite{zimmermann2019freihand} and All Camera Poses, generated as described in Sec.~\ref{subsection:AnalysisOfPose}. We use the “Procrustes aligned mean per joint position error” (PA-MPJPE), and the “mean per joint position error" (MPJPE) metrics. 
We evaluate qualitatively using both the real-world datasets and online images. We use results from online images to perform a user preference study to best assess qualitative results.

\section{Results}
\label{section:experimentsandresults}
We first evaluate both the quantitative and qualitative performance of the SHARE framework as well as its generalizability on different HPS regressors.
We then demonstrate the effectiveness of RoME sampling over other data augmentation techniques in an ablation study.

\input{tables/overall_results}
\subsection{Quantitative Results}
\label{quantitative_results}
We compare a variety of HPS models against their performance after fine-tuning with SHARE on the test datasets of 3DPW~\cite{vonMarcard20183dpw} and MPI-INF-3DHP~\cite{mono-3dhp2017}. These datasets allow us to see the performance of SHARE on diverse real-world images.

We also evaluate using our simulated evaluation datasets (``All Camera Poses'') to demonstrate the improvements overall against a wide array of camera pose variations.

As seen from Table~\ref{tab:overall_results}, SHARE improves the performance of HMR by around {\bf 30\%}, SPIN by around {\bf 20\%}, PARE by around {\bf 20\%} and CLIFF by around {\bf 20\%} when tested on the simulated datasets with all camera poses and hundreds of diverse bodies.
With the MPI-INF-3DHP and 3DPW test datasets we observed maintained performance across all baselines with improvements in some.

\subsubsection{Hand Reconstruction}
To further demonstrate that our pipeline is applicable to any HPS regressor, we extend SHARE to hand pose and shape recovery and test its performance on ExPose~\cite{choutas2020monocular} for hands.
\input{tables/hand_results}
Here we use the SHARE pipeline, with the only difference being the parametric model is that of hands, we evaluate the performance using the FreiHand test dataset~\cite{zimmermann2019freihand} as well as a simulated hand dataset, generated with from all camera poses. As seen from Table~\ref{tab:hand_results}, SHARE improves the performance of ExPose by around \textbf{20\%} when tested on All Camera Poses with some improvements on FreiHand.

\subsection{Qualitative Results}
\input{figures/qualitative}
To assess the qualitative performance of the models before and after the implementation of SHARE, we conducted a survey presenting respondents with 8 sets of 3 images. The first image, sourced online, served as the input to the model, while the second and third images represented the inferred body reconstruction using a baseline and a baseline + SHARE, both in the same color. The order of the second and third images was randomly varied. \textbf{Surveying 100 individuals, we found that respondents overwhelmingly preferred the reconstructions with SHARE, averaging 81.9\%} preference. Qualitative results are visualized in Fig.~\ref{fig:qualitative_results}, and additional results for each baseline are provided in the appendix

\subsection{Ablation Study on Sampling Techniques}
Here we compare our novel \textit{RoME} sampling technique against the greedy sampling technique (\textit{g}). We include results fine-tuned with the simple augmentation of our generated data without the SHARE sampling pipeline. The results demonstrate that the models benefit from SHARE with the \textit{RoME} sampling technique effectively improving the SHARE pipeline.

%% file: tables/overall_results.tex
\begin{table*}[ht]
    \centering
    \scalebox{0.75}{
    \begin{tabular}{ccccccc}
    \hline
         &  \multicolumn{2}{c}{3DPW~\cite{vonMarcard20183dpw}} & \multicolumn{2}{c}{MPI-INF-3DHP~\cite{mono-3dhp2017}} & \multicolumn{2}{c}{All Camera Poses}\\
       Method  & MPJPE$\downarrow$ & PA-MPJPE$\downarrow$ & MPJPE$\downarrow$ & PA-MPJPE$\downarrow$ & MPJPE$\downarrow$ & PA-MPJPE$\downarrow$
    \\
    \hline
    HMMR~\cite{kanazawa2019learning} & 116.5 & 72.6& -& -& -& -\\
    VIBE~\cite{kocabas2020vibe} & 93.5 & 56.5& 96.6& 64.6& -& -\\
    Pose2Mesh~\cite{choi2020pose2mesh} & 89.2 & 58.9& -& -& -& -\\
    I2L-MeshNet~\cite{moon2020i2l}& 93.2 & 58.6& -& -& -& -\\
    DSR~\cite{dwivedi2021learning}& 91.7 & 54.1& -& -& -& -\\
    HybrIK$^\dagger$~\cite{li2020hybrik}&80.0 &48.8 & 91.0 & - & - & - \\
    Biggs et. al~\cite{biggs20203d}& 93.8 & 59.9& -& -& -& -\\
    ProHMR~\cite{kolotouros2021probabilistic}& - & 55.1& -& 65.0& -& -\\
    Sengupta et. al~\cite{sengupta2021hierarchical}& 84.9 & 53.6& -& -& -& -\\
    HuManiFlow~\cite{sengupta2023humaniflow}& 83.9 & 53.4& -& -& -& -\\

    Doersch et al.~\cite{doersch2019sim2real}& - & 74.7& -& -& -& -\\

    MEVA~\cite{luo20203d}& 86.9 & 54.7& 96.4& 65.4& -& -\\

    LearnedGD~\cite{song2020human}&-&55.9&-&-&-&-\\
    TCMR~\cite{choi2021beyond}&95.0&55.8&97.4&62.8&-&-\\
    \hline
    HMR*~\cite{hmrKanazawa17}& 112.3 & 67.5 & 124.2& 89.8& 357.6& 154.0\\
    \textbf{HMR + SHARE}& \textbf{111.9} & \textbf{67.4}& \textbf{114.6}& \textbf{74.3}& \textbf{113.7}& \textbf{107.9}\\
    \hline
    SPIN*~\cite{kolotouros2019spin}& 96.0& 59.0& 107.1& 70.1& 364.9& 146.0\\
    \textbf{SPIN + SHARE}& 97.5 & 59.7& \textbf{104.9}& \textbf{69.8}&\textbf{138.1}&\textbf{115.7}\\
    \hline
    PARE*~\cite{kocabas2021pare}& 81.8 & 50.8& 100.12& 68.9& 327.0 & 124.3\\
    \textbf{PARE + SHARE}& \textbf{79.7} & \textbf{49.1} & \textbf{99.9}& \textbf{66.9}& \textbf{113.9}& \textbf{98.6}\\
    \hline
    CLIFF*$^\dagger$~\cite{li2022cliff}& 76.5& 48.7& 99.6&70.0 & 360.7& 135.8\\
    \textbf{CLIFF + SHARE$^\dagger$}& \textbf{74.8} & \textbf{47.30}& \textbf{98.0}&\textbf{ 67.2}& \textbf{122.8}& \textbf{108.5}\\
    \hline
    \end{tabular}
    }
    \caption{\textbf{Evaluation of SHARE and SOTA HPS on 3DPW, MPI-INF-3dHP and All Camera Poses.} All metrics are in mm. * denotes the OpenMMLab implementation of the method and $\dagger$ indicates the model has been trained with 3DPW. Fine-tuning with SHARE improves the performance of several HPS techniques (HMR, SPIN, PARE, and CLIFF) beyond its baseline capabilities and often achieve the best or comparable results across all SOTA methods.}
    \vspace*{-0.5em}
    \label{tab:overall_results}
\end{table*}

%% file: tables/hand_results.tex
\begin{table}[t]
    \centering
    \scalebox{0.9}{
    \begin{tabular}{ccc}
    \hline
         & FreiHand & All Camera Poses(Hands) \\
    Method & PA-MPJPE$\downarrow$ & PA-MPJPE$\downarrow$\\
    \hline
    Pose2Mesh~\cite{choi2020pose2mesh} & 7.40 & -\\
    I2L-MeshNet~\cite{moon2020i2l} & 7.40 & -\\
    \hline
    ExPose* (hand)~\cite{choutas2020monocular} & 10.3 & 40.0\\
    \textbf{ExPose + SHARE} &\textbf{ 9.3 }& \textbf{31.7}\\
    \hline
    \end{tabular}
    }
    \caption{\textbf{Evaluation of SHARE and SOTA HPS on FreiHand and All Camera Poses (Hands)} All metrics are in mm. * denotes the OpenMMLab implementation of the technique. Fine-tuning with SHARE improves the performance of ExPose beyond its baseline capabilities.}
    \label{tab:hand_results}
    \vspace{-1.0em}
\end{table}

%% file: figures/qualitative.tex
\begin{figure}[t]
    \centering
    \includegraphics[width=0.15\columnwidth]{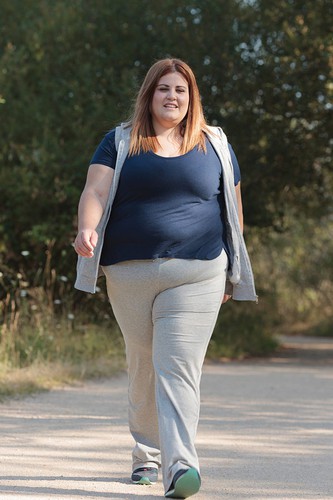}
    \includegraphics[width=0.15\columnwidth]{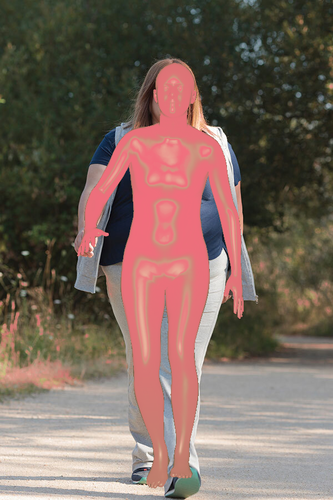}
    \includegraphics[width=0.15\columnwidth]{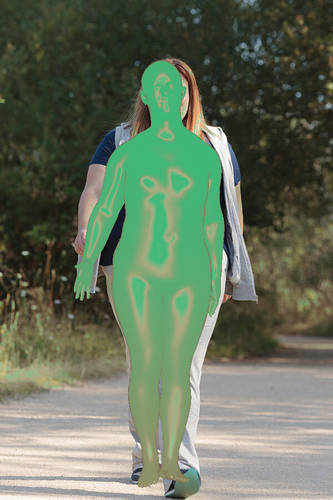}
    \includegraphics[width=0.15\columnwidth]{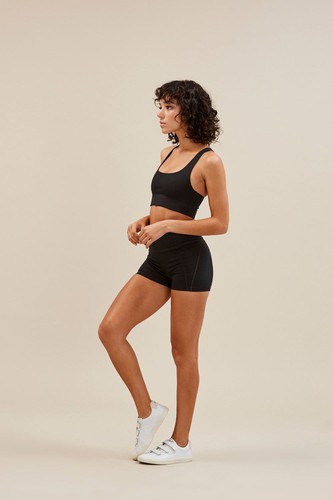}
    \includegraphics[width=0.15\columnwidth]{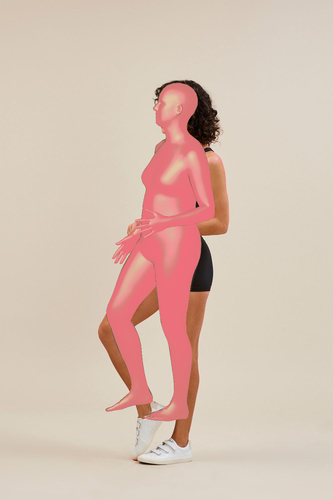}
    \includegraphics[width=0.15\columnwidth]{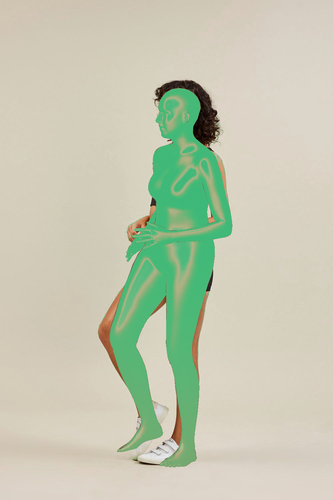}
    \includegraphics[width=0.155\columnwidth]{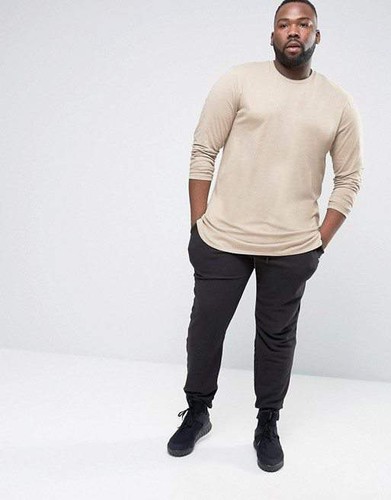}
    \includegraphics[width=0.155\columnwidth]{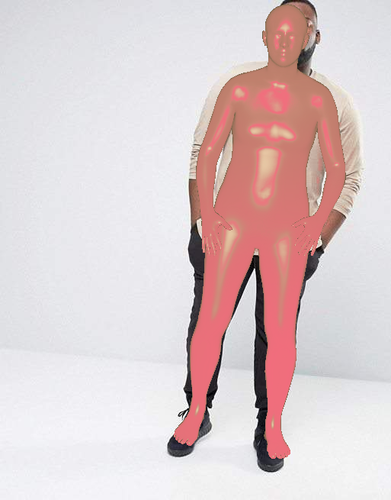}
    \includegraphics[width=0.155\columnwidth]{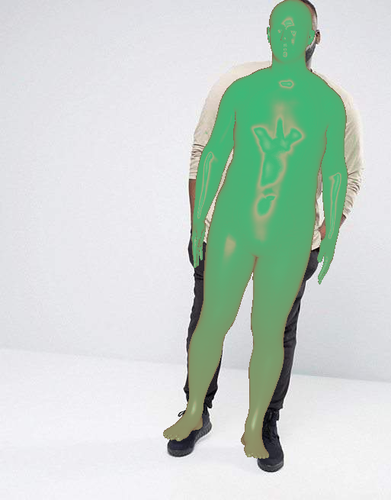}
    \includegraphics[width=0.15\columnwidth]{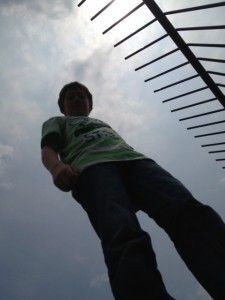}
    \includegraphics[width=0.15\columnwidth]{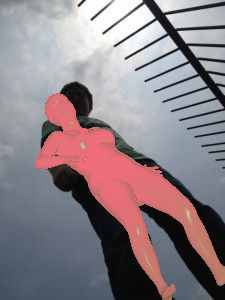}
    \includegraphics[width=0.15\columnwidth]{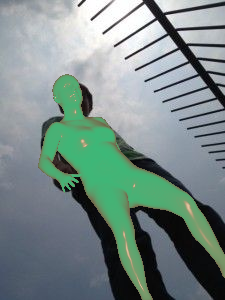}
    \includegraphics[width=0.15\columnwidth]{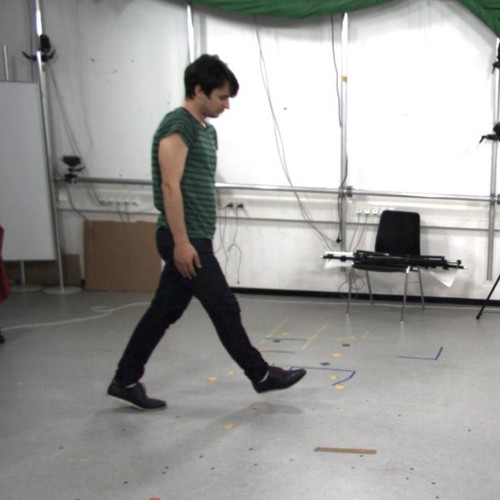}
     \includegraphics[width=0.15\columnwidth]{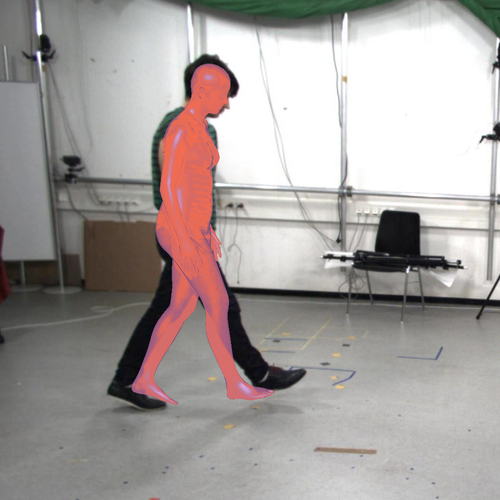}
     \includegraphics[width=0.15\columnwidth]{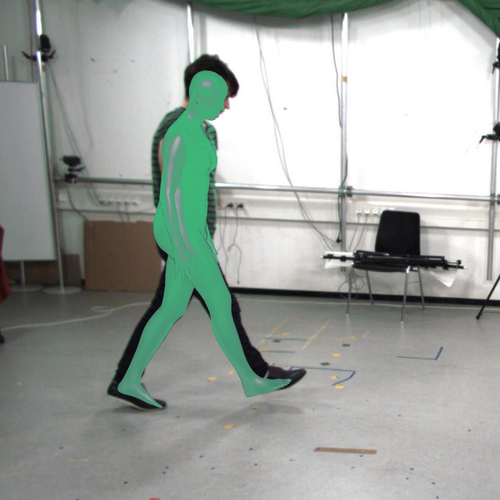}
    \includegraphics[width=0.15\columnwidth]{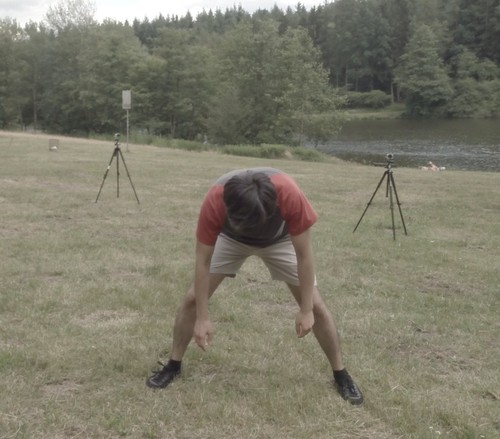}
     \includegraphics[width=0.15\columnwidth]{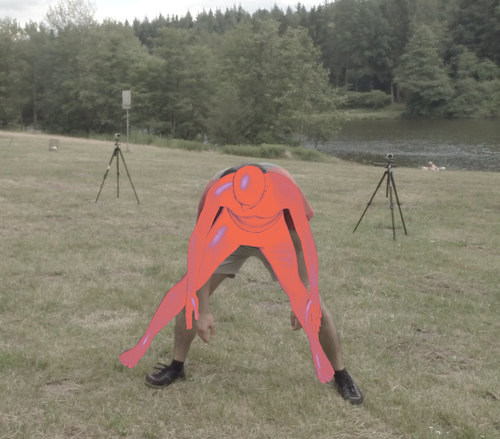}
     \includegraphics[width=0.15\columnwidth]{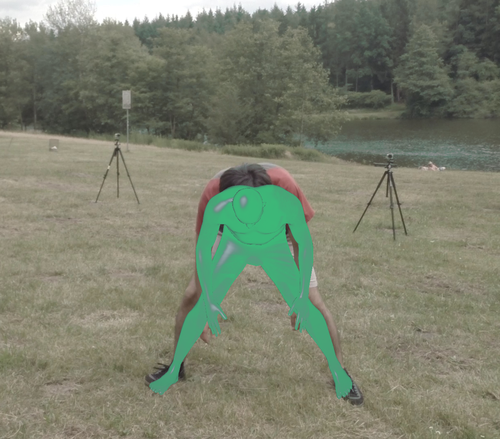}
    \includegraphics[width=0.15\columnwidth]{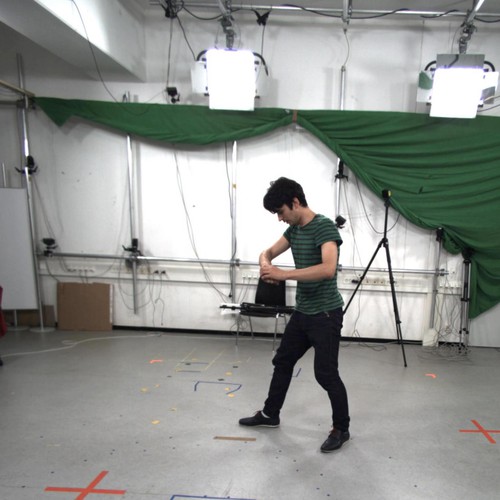}
    \includegraphics[width=0.15\columnwidth]{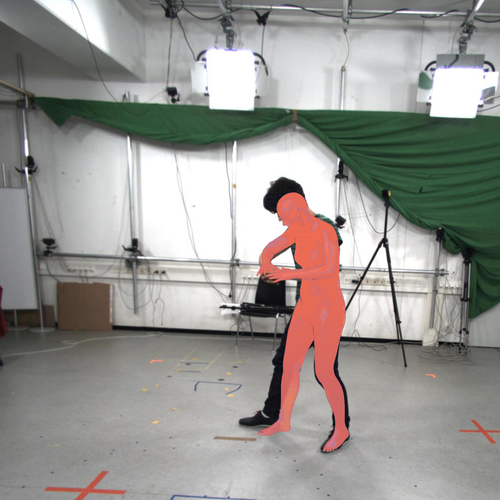}
    \includegraphics[width=0.15\columnwidth]{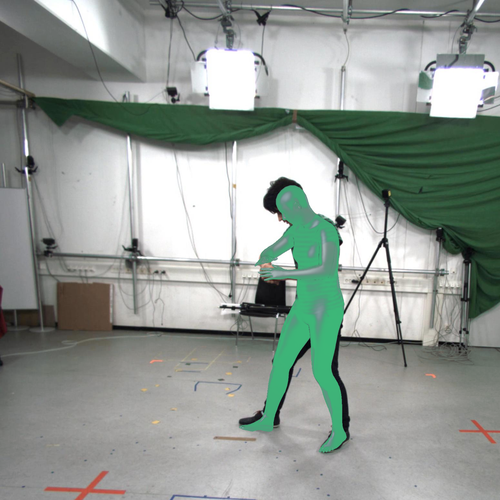}
    \includegraphics[width=0.15\columnwidth]{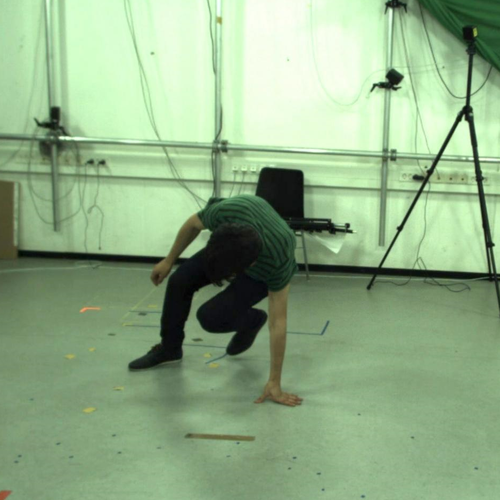}
    \includegraphics[width=0.15\columnwidth]{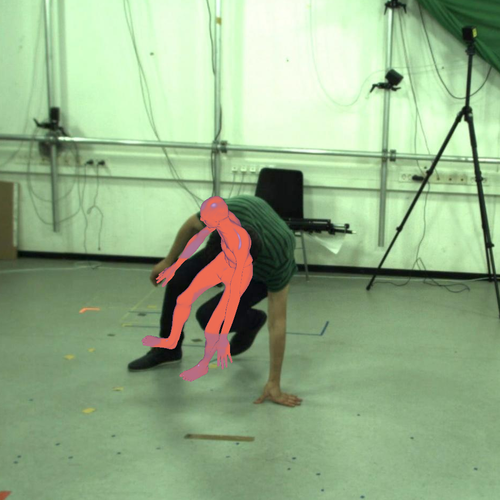}
    \includegraphics[width=0.15\columnwidth]{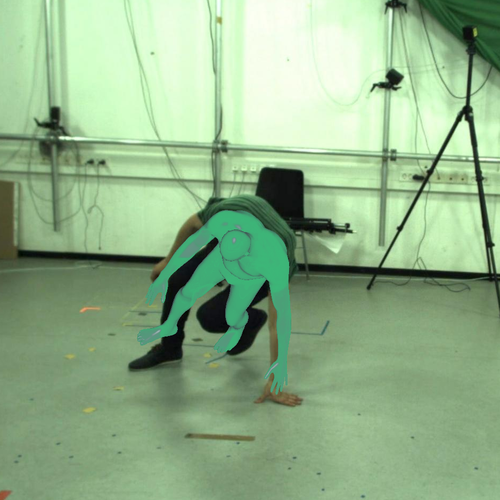}
    \vspace*{-0.5em}
    \caption{\textbf{Qualitative results on internet and MPI-INF-3DHP images using baselines~\cite{kocabas2021pare,hmrKanazawa17,li2022cliff,kolotouros2019spin} before (center in red) and after fine-tuning with SHARE (right in green).} Additional qualitative results on MPI-INF-3DHP~\cite{mono-3dhp2017} for individual baselines can be found in the appendix.}
    \vspace*{-1em}
    \label{fig:qualitative_results}
\end{figure}

%% file: sec/5_conclusion.tex
\section{Conclusion}
\input{tables/sampling_ablation_results}
In summary, our investigation highlights the susceptibility of current human pose and shape (HPS) reconstruction methods to distortions induced by varied camera poses. Our investigation into the impact of camera poses variations on reconstruction results led us to create a fine-tuning framework \textit{SHARE}, which performs dynamic adversarial data augmentation using a novel \textit{RoME} sampling technique. 
We demonstrate the performance of SHARE using various HPS methods and perform an ablation study to demonstrate the advantage gained through RoME sampling. Additionally we performed a survey to adequately evaluate the qualitative performance of SHARE.
We found that SHARE improved the baseline performances of multiple tested HPS methods and bolstered the robustness of these models against camera pose variations. We intend to publicly release the datasets and implementation of SHARE for the benefit of the research community.
\vspace{0.5em}

\noindent
{\bf Limitations and Generalization: } 
Factors such as body size, skin tones, body-environment contrast, etc. can also affect reconstruction results. SHARE holds promise for further extension to provide a more comprehensive adversarial training framework for handling other forms of image variation. 

%% file: tables/sampling_ablation_results.tex
\begin{table}[ht]
    \centering
    \scalebox{0.56}{
    \begin{tabular}{ccccccc}
    \hline
         &  \multicolumn{2}{c}{3DPW~\cite{vonMarcard20183dpw}} & \multicolumn{2}{c}{MPI-INF-3DHP~\cite{mono-3dhp2017}} & \multicolumn{2}{c}{All Camera Poses}\\
       Method  & MPJPE$\downarrow$ & PA-MPJPE$\downarrow$ & MPJPE$\downarrow$ & PA-MPJPE$\downarrow$ & MPJPE$\downarrow$ & PA-MPJPE$\downarrow$
    \\
    \hline
    
    HMR + syn. data& 132.6 & 73.7 & 122.4 & 80.8& 136.6 & 115.8\\
    HMR + SHARE (g)& 116.63 & 69.3 & 122.1& 77.8& 115.3& \textbf{99.70}\\
    \textbf{HMR + SHARE}& \textbf{111.9} & \textbf{67.4}& \textbf{114.6}& \textbf{74.3}& \textbf{113.7}& 107.9\\
    \hline
    SPIN + syn. data& 106.8& 64.2& 117.2& 73.6& 255.4& 143.7\\
    SPIN + SHARE (g)& 97.8 & 59.8& 109.4& 76.4& 159.0& 126.5\\
    \textbf{SPIN + SHARE}& \textbf{97.5 }&\textbf{ 59.7}& \textbf{104.9}& \textbf{69.8}&\textbf{138.1}& \textbf{115.7}\\
    \hline
    PARE + syn. data& 84.6 & 50.3 & 104.9& 69.6& 115.1& 102.2\\
    PARE + SHARE (g)& 88.5 & 51.4& 100.6& 67.5& 117.6& 100.7\\
    \textbf{PARE + SHARE}& \textbf{79.7} & \textbf{49.1} & \textbf{99.9}& \textbf{66.9}& 113.9& 98.6\\
    \hline
    CLIFF + syn. data& 81.9 & 52.7& 111.8& 73.5& 147.7& 116.6\\
    CLIFF + SHARE (g)& 80.6 & 50.7& 101.4& 70.2& 143.4& 113.6\\
    \textbf{CLIFF + SHARE$^\dagger$}& \textbf{74.8} & \textbf{47.30}& \textbf{98.0}&\textbf{ 67.2}& \textbf{122.8}& \textbf{108.5}\\
    \hline
    \end{tabular}
    }
    \caption{\textbf{Ablation study of Sampling techniques in SHARE on 3DPW, MPI-INF-3dHP and All Camera Poses.} All metrics are in mm. $\dagger$ indicates the model has been trained with 3DPW and (g) indicates SHARE with \textit{greedy} sampling. \textit{RoME} sampling offers improvements over greedy sampling and the original techniques (HMR, SPIN, PARE, and CLIFF).}
    \vspace*{-1em}
    \label{tab:ablation}
\end{table}

%% file: sec/X_suppl.tex
\clearpage

\setcounter{page}{1}
\maketitlesupplementary

\section{Qualitative Results}
Figures~\ref{fig:hmr_qual},~\ref{fig:spin_qual} and ~\ref{fig:pare_qual} visualize qualitative results of each baseline before and after SHARE.

\section{Sensitivity Analysis on HPS techniques with Respect to Camera Pose}
Fig~\ref{fig:all_cyclical} plots for the sensitivity analyses for all baselines with respect to camera poses. With each technique we see a discernible oscillatory bias with varying performance in different regions around the human body, the mean and variance of the errors drops significantly after the implementation of SHARE.
\input{figures/cyclical_all}

\input{figures/all_mesh}

\section{Additional Implementation Details}
For our adversarial training, we augmented 15\% of the original training data at every iteration.
Using this configuration we perform fine-tuning on the baseline HPS versions.

For each cycle of SHARE we sample 50,000 images from our synthetic training dataset. One training interval of SHARE consists of 5 epochs. Our synthetic testing dataset is composed of 10,000 images. Each dataset contains images from 2500 camera poses. For each image, we sampled
diverse realistic bodies, poses, skin tones, clothing, lighting, and environments. We trained our model on multiple servers: a dual NVIDIA 3090 machine takes around 8 hours for 40 epochs.

\input{figures/hmr_qual}
\input{figures/spin_qual}
\input{figures/pare_qual}
\input{figures/cliff_qual}

%% file: figures/cyclical_all.tex
\begin{figure}
    \centering
    \includegraphics[width=\columnwidth]{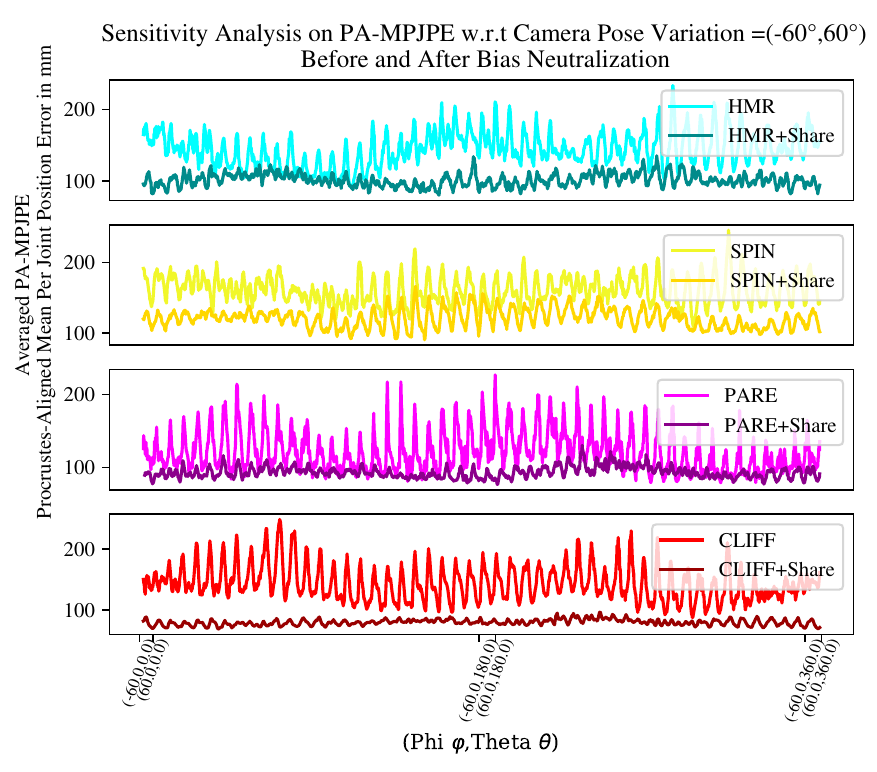}
    \caption{\textbf{Sensitivity analysis on HMR~\cite{hmrKanazawa17}, SPIN~\cite{kolotouros2019spin}, PARE~\cite{kocabas2021pare} and CLIFF~\cite{li2022cliff} before and after SHARE with respect to camera pose using PA-MPJPE.} The x-axis iterates through all camera poses $(\theta,\phi)$, where $\phi$ represents the azimuthal angle around the body (0, 360), and $\theta$ represents the vertical viewing angle (-60, 60) for each $\phi$. The y-axis represents the average error in PA-MPJPE over a diverse dataset encompassing a wide range of bodies, body poses, and environments. This plot explicitly depicts the average error associated with each camera pose, revealing \textbf{a discernible oscillatory bias with varying performance in different regions around the human body, with a significant decrease in variance with SHARE}}
    \label{fig:all_cyclical}
\end{figure}

%% file: figures/all_mesh.tex
\begin{figure}[t]
    \centering
    \includegraphics[width=0.23\textwidth]{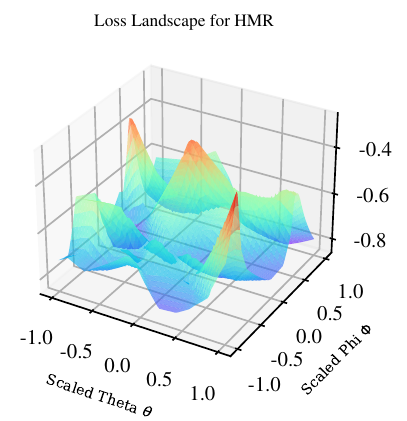}
    \includegraphics[width=0.23\textwidth]{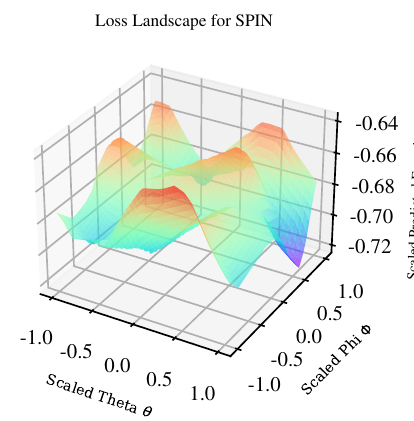}
    \newline
    \includegraphics[width=0.23\textwidth]{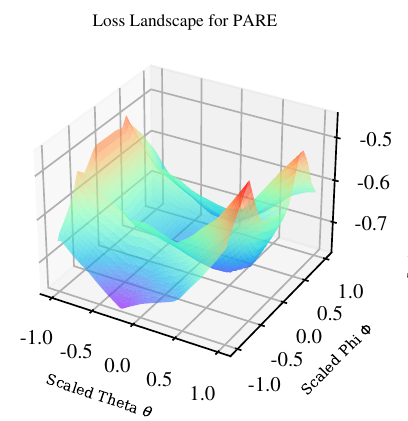}
    \includegraphics[width=0.23\textwidth]{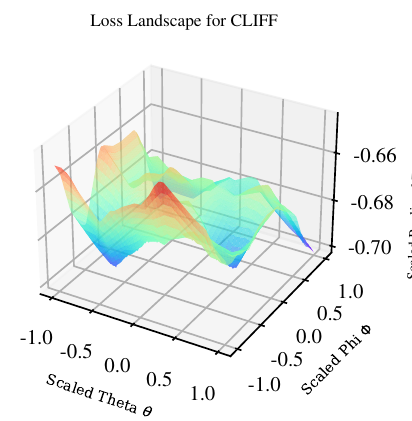}

\caption{Loss Landscapes for All Baseline models}
\label{fig:all_landscapes}
\end{figure} 

%% file: figures/hmr_qual.tex
\begin{figure*}[h]

    
    \centering
    \includegraphics[width=0.3\textwidth]{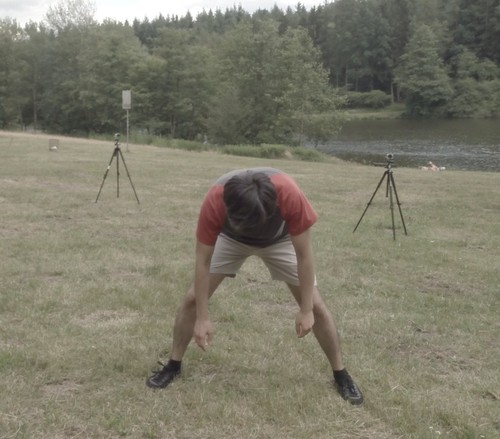}
    \includegraphics[width=0.3\textwidth]{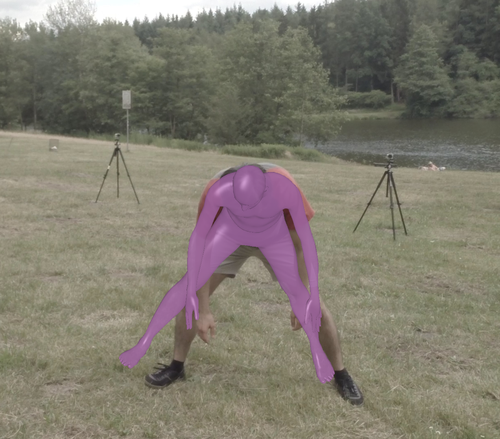}
    \includegraphics[width=0.3\textwidth]{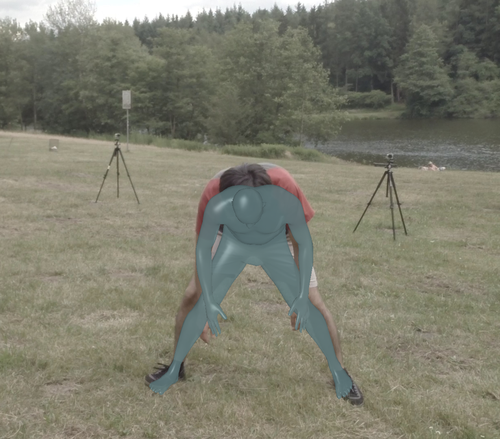}
    \includegraphics[width=0.3\textwidth]{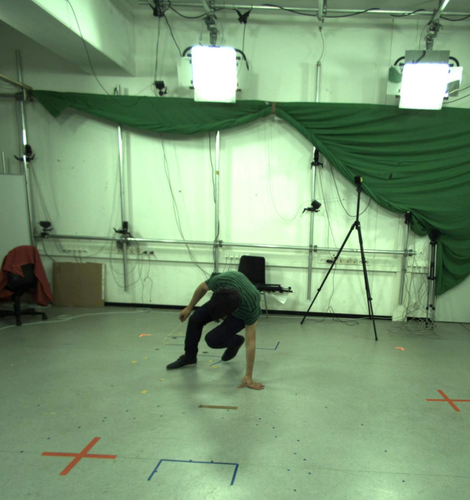}
    \includegraphics[width=0.3\textwidth]{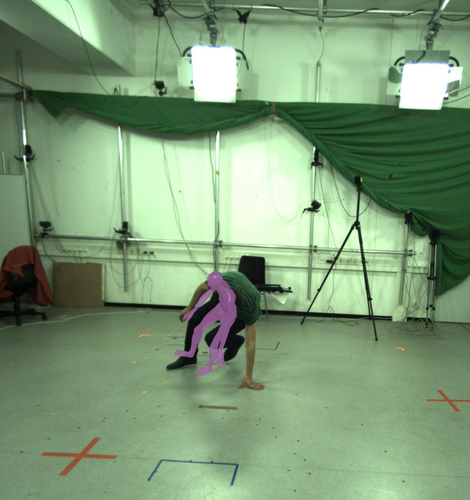}
    \includegraphics[width=0.3\textwidth]{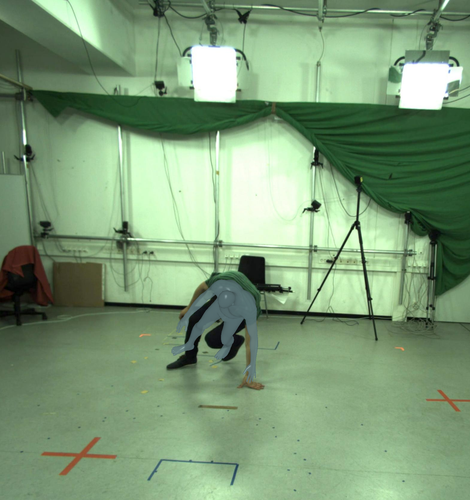}
    \includegraphics[width=0.3\textwidth]{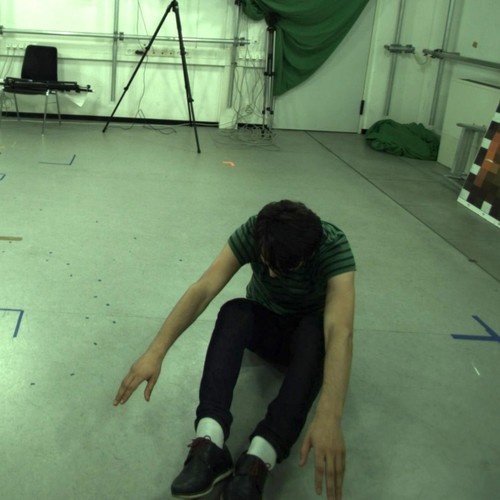}
    \includegraphics[width=0.3\textwidth]{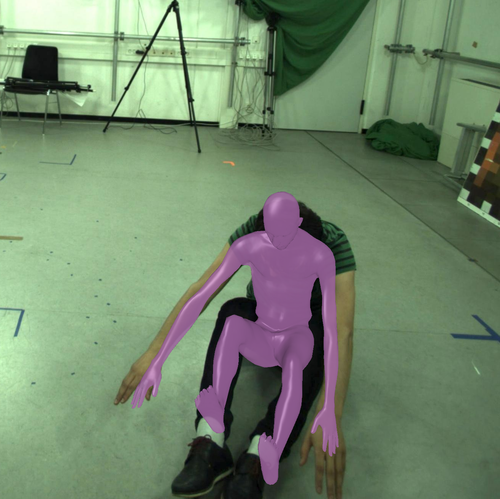}
    \includegraphics[width=0.3\textwidth]{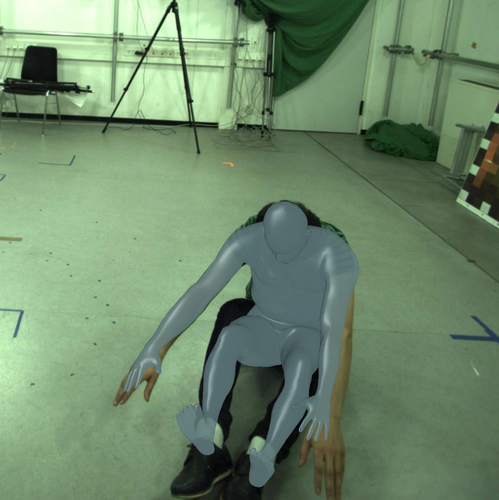}

\caption{Qualitative results on reference images~\cite{mehta2017monocular,vonMarcard20183dpw} (left) of HMR~\cite{hmrKanazawa17} (center) and HMR + SHARE (right).}
\label{fig:hmr_qual}
\end{figure*}

%% file: figures/spin_qual.tex
\begin{figure*}[h]
    \centering

   
    \includegraphics[width=0.3\textwidth]{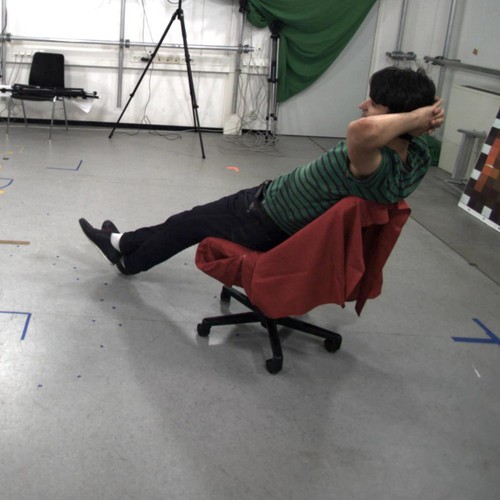}
    \includegraphics[width=0.3\textwidth]{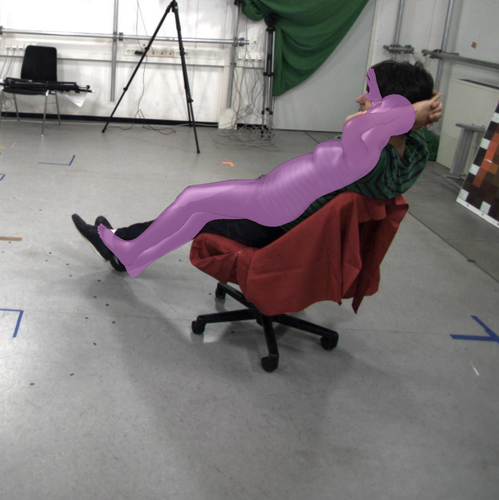}
    \includegraphics[width=0.3\textwidth]{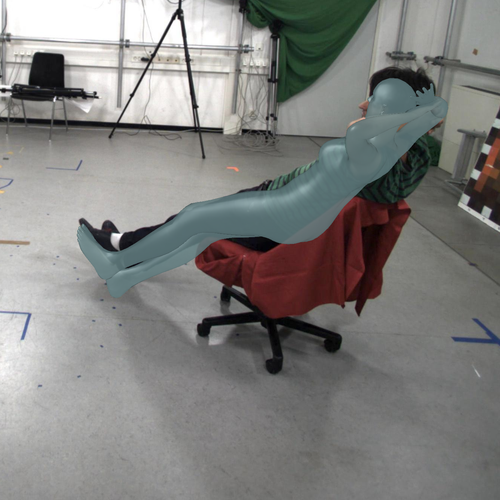}
    \includegraphics[width=0.3\textwidth]{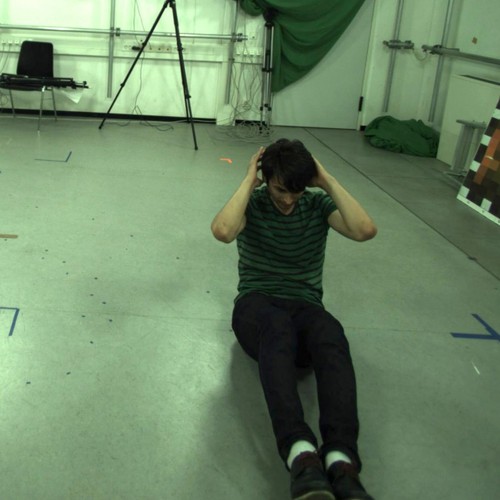}
    \includegraphics[width=0.3\textwidth]{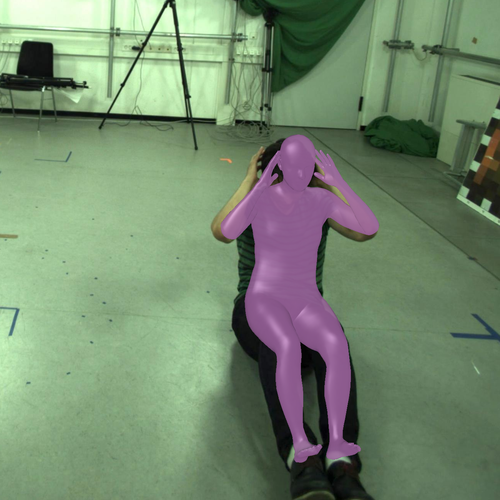}
    \includegraphics[width=0.3\textwidth]{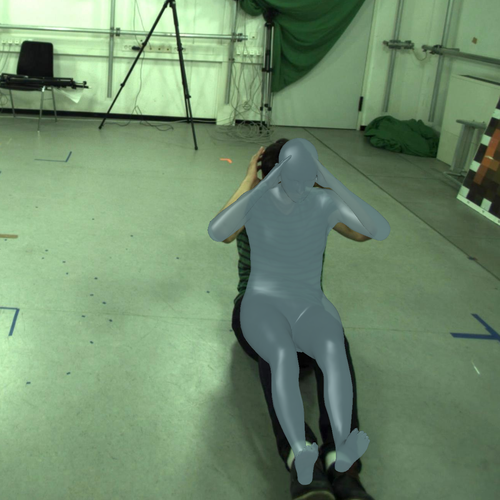}
    \includegraphics[width=0.3\textwidth]{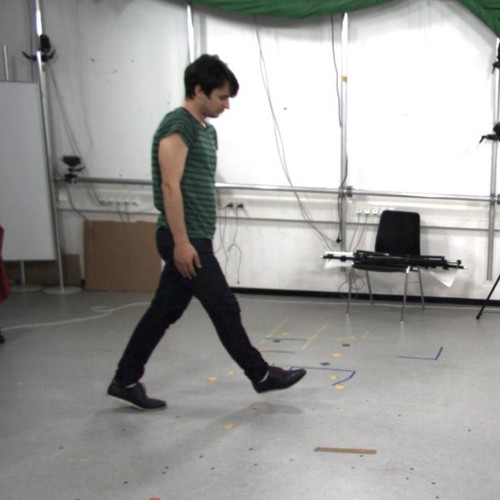}
    \includegraphics[width=0.3\textwidth]{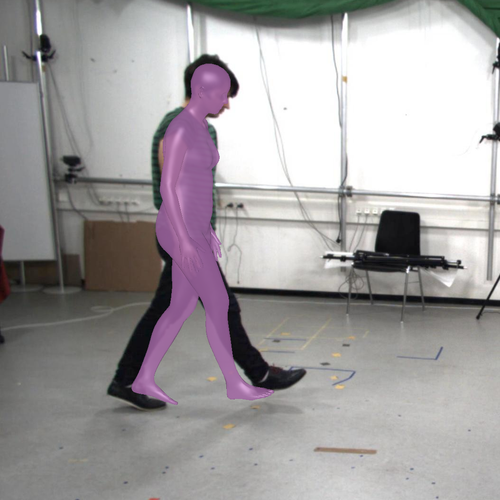}
    \includegraphics[width=0.3\textwidth]{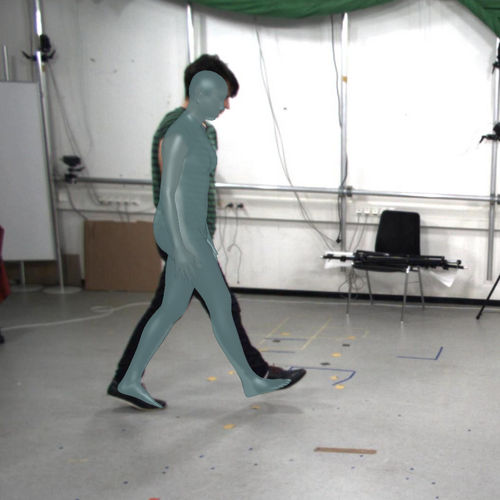}
    
\caption{Qualitative results on reference images~\cite{mehta2017monocular} (left) of SPIN~\cite{kolotouros2019spin} (center) and SPIN + SHARE (right).}
\label{fig:spin_qual}
\end{figure*} 

%% file: figures/pare_qual.tex
\begin{figure*}[h]
    \centering

   
    \includegraphics[width=0.3\textwidth]{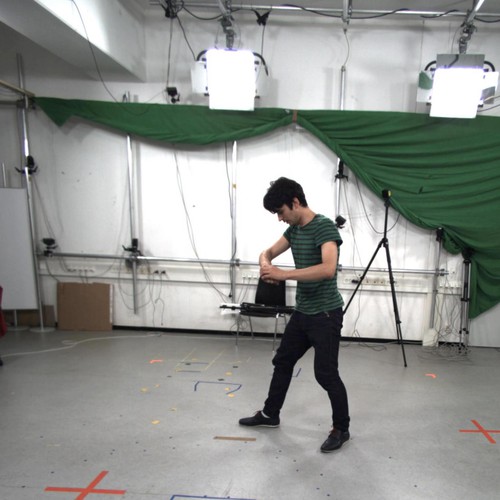}
    \includegraphics[width=0.3\textwidth]{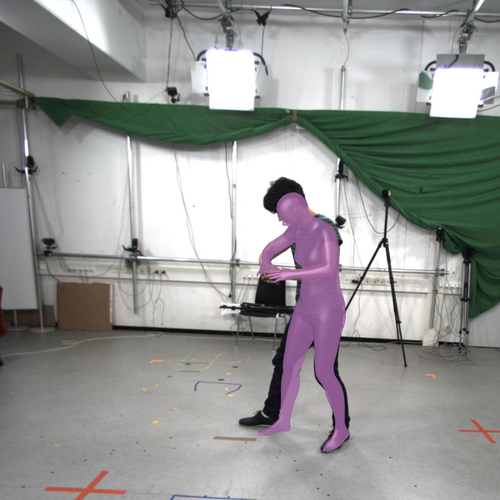}
    \includegraphics[width=0.3\textwidth]{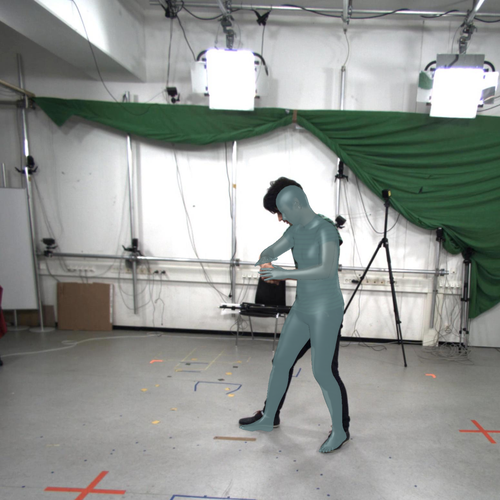}
    \includegraphics[width=0.3\textwidth]{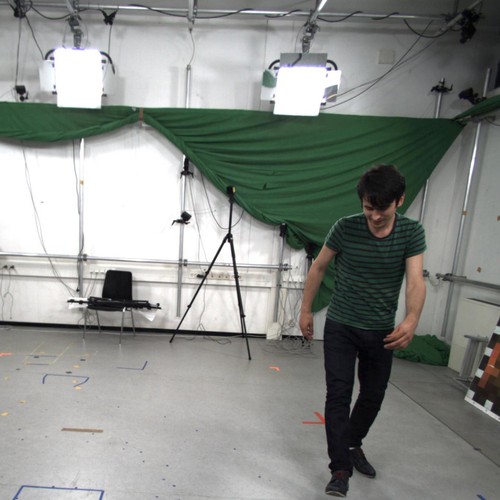}
    \includegraphics[width=0.3\textwidth]{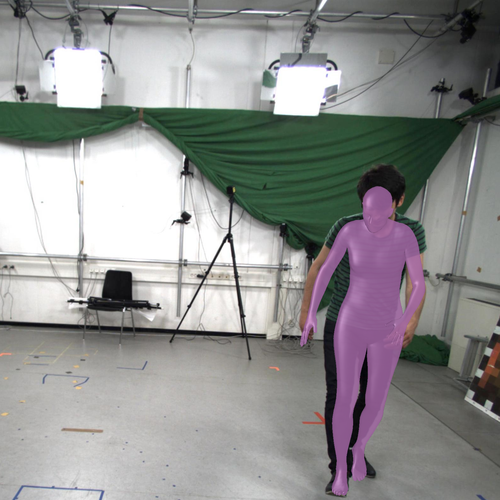}
    \includegraphics[width=0.3\textwidth]{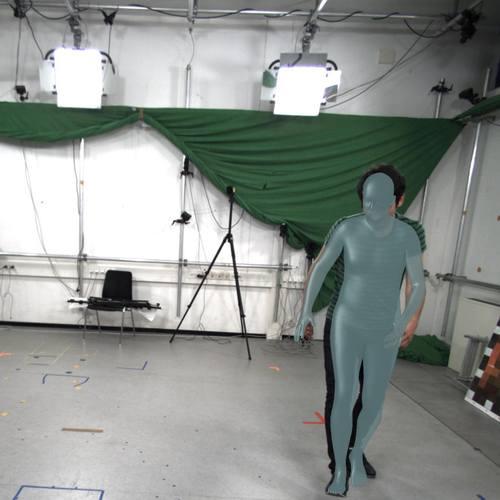}
    \includegraphics[width=0.3\textwidth]{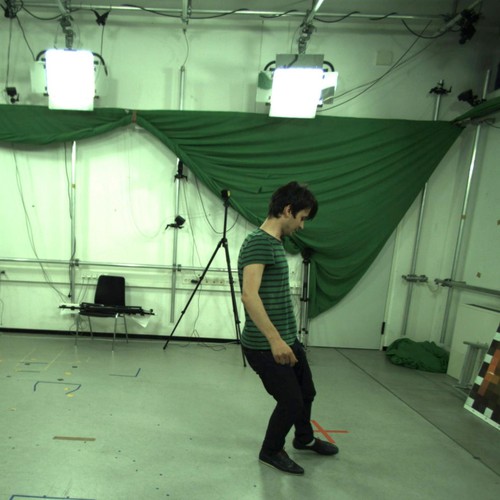}
    \includegraphics[width=0.3\textwidth]{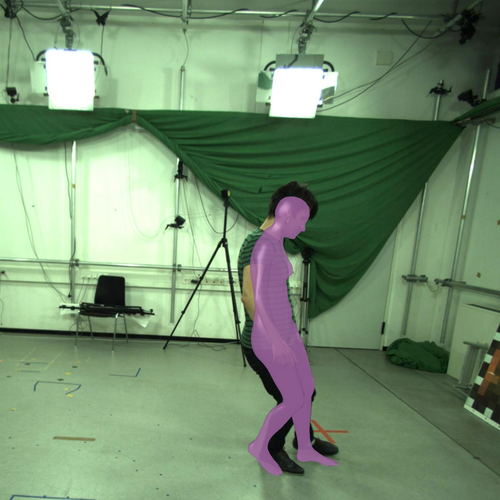}
    \includegraphics[width=0.3\textwidth]{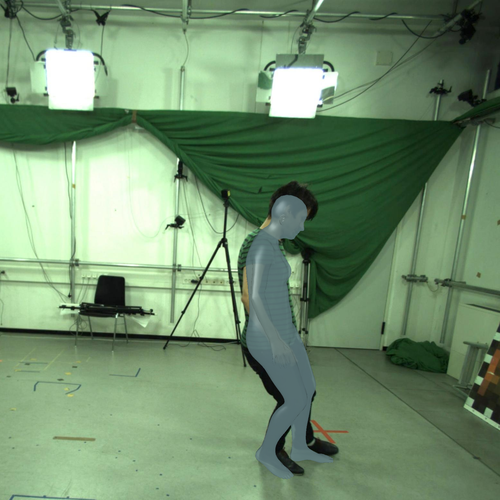}
    
\caption{Qualitative results on reference images~\cite{mehta2017monocular} (left) of PARE~\cite{kocabas2021pare} (center) and PARE + SHARE (right).}
\label{fig:pare_qual}
\end{figure*} 

%% file: figures/cliff_qual.tex
\begin{figure*}[h]
    \centering

   
    \includegraphics[width=0.3\textwidth]{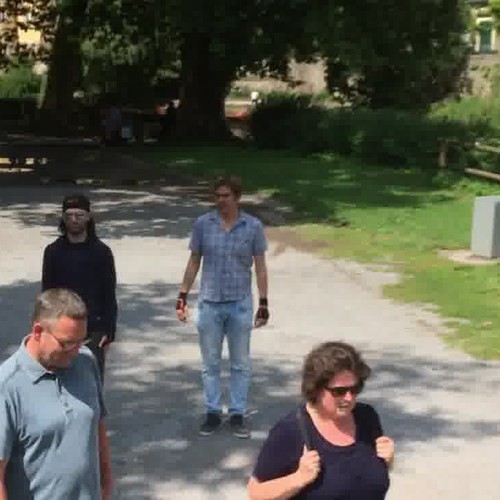}
    \includegraphics[width=0.3\textwidth]{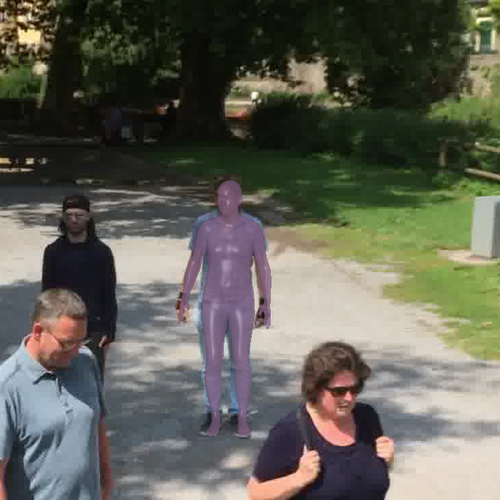}
    \includegraphics[width=0.3\textwidth]{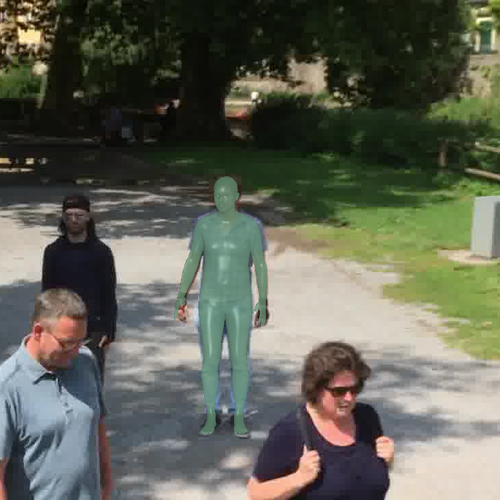}
    \includegraphics[width=0.3\textwidth]{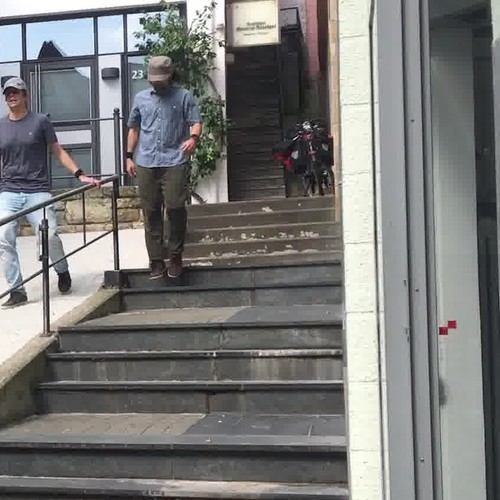}
    \includegraphics[width=0.3\textwidth]{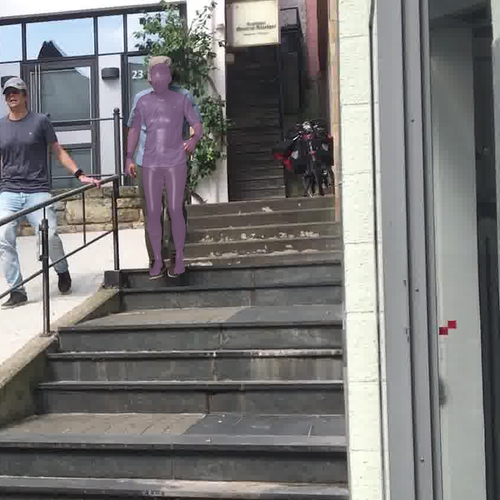}
    \includegraphics[width=0.3\textwidth]{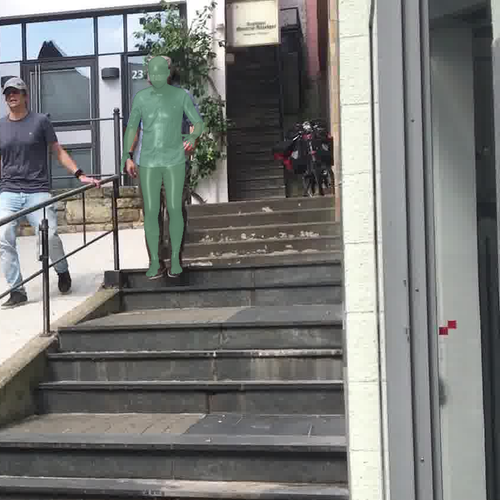}
    \includegraphics[width=0.3\textwidth]{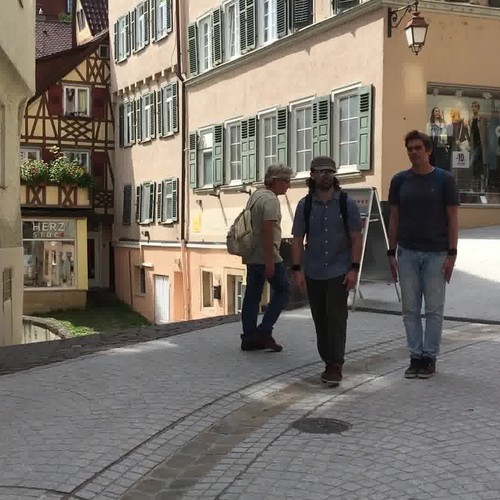}
    \includegraphics[width=0.3\textwidth]{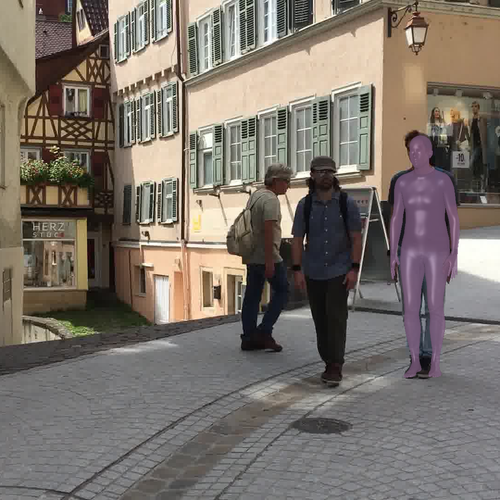}
    \includegraphics[width=0.3\textwidth]{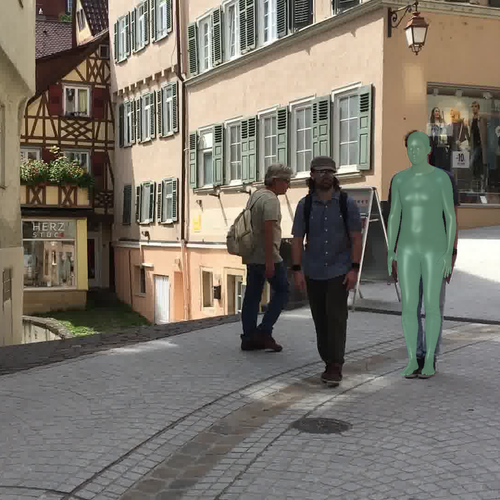}z
    
\caption{Qualitative results on reference images~\cite{vonMarcard20183dpw} (left) of CLIFF~\cite{li2022cliff} (center) and CLIFF + SHARE (right).}
\label{fig:cliff_qual}
\end{figure*}